\documentclass[lettersize,journal]{IEEEtran}
\usepackage{amsmath,amssymb,amsfonts}
\usepackage{utfsym}
\usepackage{algorithmic}
\usepackage{algorithm}
\usepackage{array}

\usepackage{textcomp}
\usepackage{stfloats}
\usepackage{url}
\usepackage{verbatim}
\usepackage{graphicx}
\usepackage{cite}
\usepackage{booktabs}
\usepackage{amsthm}
\usepackage{color}
\usepackage{xcolor}
\usepackage{soul}
\usepackage{multirow}
\usepackage{boldline} 
\usepackage{threeparttable}
\usepackage{float}
\usepackage[colorlinks=true]{hyperref}

\usepackage{makecell}

\usepackage{xcolor} 

\usepackage{graphicx} 
\usepackage{subcaption}

\graphicspath{{figures/}}
\hypersetup{
hidelinks,
colorlinks=true,
linkcolor=black,
citecolor=black
}

\begin{document}

\title{A Benchmark for Multi-Lingual Vision-Language Learning in Remote Sensing Image Captioning}

\author{Qing~Zhou,
        Tao~Yang,
        Junyu~Gao,~\IEEEmembership{Member,~IEEE},
        Weiping~Ni,
        Junzheng~Wu, and
        Qi~Wang,~\IEEEmembership{Senior Member,~IEEE}

\IEEEcompsocitemizethanks{

Qing Zhou, Tao Yang, Junyu Gao, and Qi Wang are with the School of Artificial Intelligence, Optics and Electronics (iOPEN), Northwestern Polytechnical University, Xi'an 710072, Shaanxi, P. R. China (e-mail: \href{mailto:chautsing@gmail.com}{\textcolor{black}{chautsing@gmail.com}}, \href{mailto:taoytao@mail.nwpu.edu.cn}{\textcolor{black}{taoytao@mail.nwpu.edu.cn}}, \href{mailto: gjy3035@gmail.com}{\textcolor{black}{ gjy3035@gmail.com}}, \href{mailto:crabwq@gmail.com}{\textcolor{black}{crabwq@gmail.com}}).

Weiping Ni and Junzheng Wu are with the Department of Remote Sensing, Northwest Institute of Nuclear Technology, Xi'an 710024, Shaanxi, P. R. China (e-mail: \href{mailto:niweiping@nint.ac.cn}{\textcolor{black}{niweiping@nint.ac.cn}}, \href{mailto:wujunzheng@nint.ac.cn}{\textcolor{black}{wujunzheng@nint.ac.cn}}).

Qing Zhou and Tao Yang are equal contribution.

Qi Wang is the corresponding author.
}
}

\maketitle

\begin{abstract}

Remote Sensing Image Captioning (RSIC) is a cross-modal field bridging vision and language, aimed at automatically generating natural language descriptions of features and scenes in remote sensing imagery. Despite significant advances in developing sophisticated methods and large-scale datasets for training vision-language models (VLMs), two critical challenges persist: the scarcity of non-English descriptive datasets and the lack of multilingual capability evaluation for models. These limitations fundamentally impede the progress and practical deployment of RSIC, particularly in the era of large VLMs.
To address these challenges, this paper presents several significant contributions to the field.
First, we introduce and analyze BRSIC (Bilingual Remote Sensing Image Captioning), a comprehensive bilingual dataset that enriches three established English RSIC datasets with Chinese descriptions, encompassing 13,634 images paired with 68,170 bilingual captions.
Building upon this foundation, we develop a systematic evaluation framework that addresses the prevalent inconsistency in evaluation protocols, enabling rigorous assessment of model performance through standardized retraining procedures on BRSIC.
Furthermore, we present an extensive empirical study of eight state-of-the-art large vision-language models (LVLMs), examining their capabilities across multiple paradigms including zero-shot inference, supervised fine-tuning, and multi-lingual training. This comprehensive evaluation provides crucial insights into the strengths and limitations of current LVLMs in handling multilingual remote sensing tasks.
Additionally, our cross-dataset transfer experiments reveal interesting findings. While traditional models show superior performance when transferring between visually similar datasets, LVLMs demonstrate more robust and balanced capabilities across different dataset scales and languages. These findings provide useful insights for advancing multilingual RSIC research.
The code and data will be available at \url{https://github.com/mrazhou/BRSIC}.

\end{abstract}

\section{Introduction}
\IEEEPARstart{R}{emote} sensing image captioning (RSIC) aims to describe the visual information in remote sensing images (e.g., object states, quantities, location distributions, and scene attributes) in a human-understandable natural language, which is a visual-language multimodal task. Based on these generated text descriptions, non-experts can more easily understand remote sensing images and provide useful references for disaster monitoring and agricultural management~\cite{4215085,sen,885202,9308980,9153154}. To achieve better RSIC performance, advanced models and large-scale high-quality datasets are constantly proposed and constructed. In terms of methods, deep learning methods based on sequence generation have evolved from local modeling-oriented CNNs and LSTMs~\cite{jiang2018learning, yao2017boosting, qu2016deep} to long-distance modeling-oriented Transformers and Attention~\cite{mlat,MG,bui2023uit,aoanet,lu2019sound,zhang2019description, zhang2019lam, liu2022remote}, and the latest Mamba architecture~\cite{Rscama, gu2023mamba} with linear complexity for long-distance modeling. Particularly, with the excellent comprehensive capabilities and zero-shot generalization of large language models (LLMs)~\cite{bai2023qwen, cai2024internlm2,llama2}, large vision-language models (LVLMs)~\cite{wang2024qwen2,internvl2,internvl2.5,lu2024deepseek,llava} also develop rapidly and are applied to image captioning.
In terms of datasets, the two small-scale datasets with a land use theme, UCM and Sydney~\cite{qu2016deep}, contain 21 scene categories and 2,100 images, and 7 scene categories and 613 images, respectively. Later, RSICD~\cite{lu2017exploring} with 30 scene categories and 10,921 images is released, and recently, NWPU-Captions~\cite{nwpu-caption} with 31,500 images covering 45 categories is released. Li et al.~\cite{li2020multi} finds that UCM, Sydney, and RSICD have many spelling and grammar errors and corrected them. Overall, the datasets are developing towards larger scale, more scenes, and higher quality.

\begin{figure}[t]
    \centering
    \includegraphics[width=0.5\textwidth]{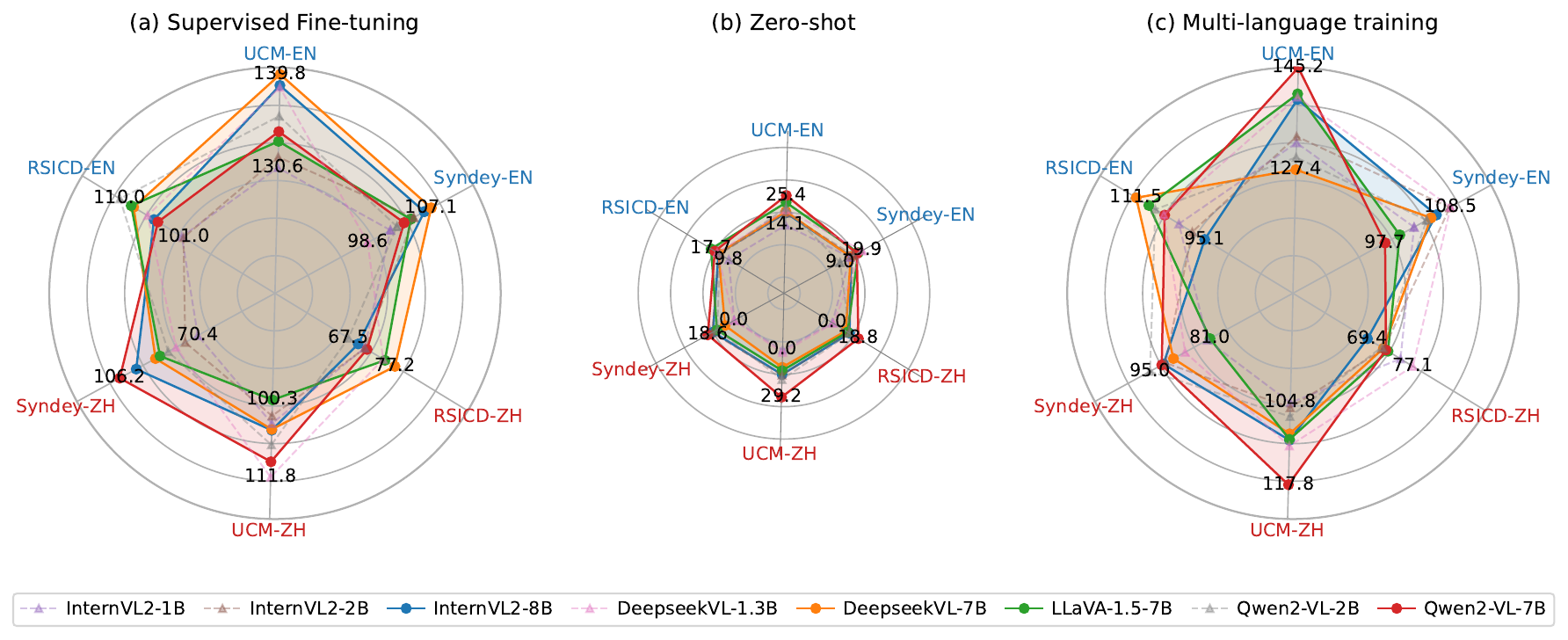}
    \caption{Performance ($(B@4+M+R+C)/4$) comparison of supervised fine-tuning, zero-shot inference and multi-lingual training on BRSIC for state-of-the-art LVLMs}
    \label{fig:1_intro}
\end{figure}

However, two key issues have been overlooked: \textbf{\textit{linguistic diversity of datasets}} and \textbf{\textit{multi-lingual adaptability of models}}.
First, \textit{linguistic diversity} in datasets is crucial for vision-language learning. Natural image captioning, initially based on English annotations, has expanded to include versions in Chinese, Japanese, and other languages, fostering significant advancements in multilingual captioning models~\cite{lan2017fluency,li2019coco,hitschler2016multimodal,pham2024ktvic}. This evolution has catalyzed the development of sophisticated multilingual vision-language large models. Despite the well-established significance of multilinguality in natural image domains, existing RSIC datasets remain exclusively English-centric, creating a substantial gap in multilingual capabilities.

Second, \textit{multi-lingual adaptability} represents the performance of models on different language datasets, whether the model architecture is language-independent.
While natural image domains have demonstrated that multilingual data integration and architectural innovations can yield models with robust multi-lingual performance~\cite{li2019coco,DBLP:conf/cvpr/ZhouZW0LYL21,Tsutsui2017UsingAT,Xiao2020AnIM}, this capability remains largely unexplored in remote sensing contexts. Although current models excel in English-based tasks, their effectiveness across linguistic boundaries remains uncertain.
The absence of multilingual annotations coupled with inconsistent evaluation protocols has created significant methodological challenges. A notable example is the RSICD dataset modifications, where researchers~\cite{li2020multi} implemented both sentence corrections and swapping of validation and test sets, leading to incomparable performance metrics across studies. This inconsistency undermines systematic evaluation and impedes the development of multilingual solutions. The current practice of transferring high-performing English-language models to other languages, while intuitively appealing, lacks empirical validation.

To address the first issue, \textit{linguistic diversity}, we construct a Chinese dataset, RSICN, based on three widely used English-annotated RSIC datasets: UCM, Sydney, and RSICD. Together with their original English counterparts, they form the BRSIC dataset, comprising a total of 13,634 images and 68,170 bilingual caption sentences. The construction of the RSICN dataset involved leveraging machine translation to generate initial Chinese descriptions, followed by manual correction to eliminate translation errors, ensuring data quality while significantly reducing the time and effort required for manual annotation.
Subsequently, we conduct a comprehensive comparative analysis of the English and Chinese datasets in terms of data distributions, word frequency distributions, vocabulary size, object distribution and part of speech distribution. The results reveal key insights: 
1) Uneven distribution of text features is observed across the training, validation, and test splits of the RSICD dataset. This imbalance led to improved evaluation metrics when Li et al.~\cite{li2020multi} swaps the test and validation sets. However, subsequent studies did not specify whether such a swap was applied, resulting in inconsistent evaluations and a lack of comparability across methods.
2) Differences in vocabulary size and word frequency between English and Chinese datasets contributed to performance disparities. Specifically, the Chinese vocabulary is significantly larger, with fewer occurrences per word, which introduces greater modeling complexity and results in lower captioning performance in Chinese compared to English.

To address the second issue, \textit{multi-lingual adaptability}, we design a comprehensive evaluation framework addressing both methodological inconsistencies and multi-lingual challenges. We first establish a controlled testbed by retraining representative RSIC models under identical conditions across English, modified English, and Chinese variants of BRSIC. This standardized evaluation eliminates the ambiguity in previous studies caused by inconsistent dataset configurations. Furthermore, we conduct the first comprehensive assessment of current LVLMs through three evaluation settings: zero-shot inference to assess their inherent multi-lingual capabilities, supervised fine-tuning to explore their adaptation potential, and multi-lingual training to improve their multi-lingual performance. 
Additionally, we conduct cross-dataset transfer to examine their generalization across different remote sensing contexts.
As illustrated in Fig.~\ref{fig:1_intro}, our extensive experiments reveal that supervised fine-tuning consistently outperforms other approaches across all datasets, demonstrating strong performance particularly in English captions. However, zero-shot inference exhibits substantially lower performance, revealing significant limitations in multi-lingual generalization capabilities. Notably, multi-lingual training emerges as a promising middle-ground approach, achieving intermediate performance levels while maintaining consistency across different languages without requiring language-specific fine-tuning. These findings underscore that despite their success in natural image tasks, existing LVLMs still face considerable challenges in effectively handling multi-lingual remote sensing scenarios, particularly in zero-shot settings.

The contributions can be summarized as follows:
\begin{itemize}
\item Dataset Construction: We develop RSICN, a comprehensive Chinese-annotated dataset that complements three widely-adopted English RSIC datasets. The resulting BRSIC dataset, featuring parallel annotations in both languages, provides essential resources for multi-lingual RSIC research. To best of our knowledge, this is the first dataset that contains parallel annotations in both English and Chinese for RSIC.
\item Benchmark Establishment: We create a standardized evaluation framework by systematically retraining and assessing existing RSIC methods under uniform conditions. This framework addresses the prevalent issue of inconsistent evaluation protocols in previous studies and enables fair performance comparisons across language settings.

\item LVLMs Evaluation: We conduct extensive experiments on the BRSIC dataset, evaluating the performance of 8 state-of-the-art LVLMs through zero-shot inference, supervised fine-tuning, and multi-lingual training. Additionally, we compare these models with traditional methods under cross-dataset transfer settings to analyze their generalization capabilities.
\end{itemize}

\section{Related Work}

\subsection{Multi-Lingual Natural Image Captioning}
Multi-lingual image captioning has emerged as a important research direction in computer vision and natural language processing. Early efforts primarily focus on English-centric datasets and models~\cite{vinyals2015show}, with subsequent works extending to multiple languages through various approaches. This evolution reflects the growing need for inclusive AI systems that can serve diverse linguistic communities.
The development of multi-lingual datasets has been fundamental to this field. Notable datasets include MS-COCO with Japanese annotations (STAIR Captions~\cite{yoshikawa2017stair}), German translations (Multi30k~\cite{elliott2016multi30k}), and Chinese versions (COCO-CN~\cite{li2019coco}). These datasets provide diverse linguistic perspectives for the same visual content. Recent additions like UIT-OpenViIC~\cite{bui2023uit} and KTVIC~\cite{pham2024ktvic} have further expanded the language coverage to Vietnamese, demonstrating increasing attention to Asian languages. Each dataset brings unique characteristics: STAIR Captions offers extensive Japanese annotations, Multi30k focuses on high-quality German translations, while COCO-CN provides culturally adapted Chinese captions.
Methodologically, various approaches have been explored for multi/cross-lingual image captioning. Early works like Miyazaki and Shimizu \cite{miyazaki2016cross} explore the possibility of generating captions directly in target languages. Some studies investigate hybrid approaches that leverage both visual features and machine translation capabilities \cite{hitschler2016multimodal}. Lan et al. \cite{lan2017fluency} propose a fluency-guided approach that focuses on generating more natural descriptions in target languages. These methods have evolved to better handle cultural nuances and linguistic characteristics of target languages, though they typically require substantial parallel data and computational resources. Current research trends focus more on end-to-end approaches that can directly generate captions in multiple languages while preserving cultural context and linguistic nuances.

\subsection{Remote Sensing Image Captioning}
Remote sensing image captioning presents unique challenges compared to natural image captioning due to the distinctive characteristics of remote sensing imagery (RSI). These images often contain complex spatial relationships, multiple objects viewed from above, and specialized domain knowledge requirements. The bird's-eye perspective introduces additional complexity in object recognition and relationship description, requiring specialized approaches different from natural image captioning.

The methodological evolution in RSIC has progressed through three main stages, driven by the unique challenge of simultaneously identifying objects and understanding their spatial relationships in RSI.
The first stage begins with CNN-RNN frameworks, adapting successful architectures from natural image captioning~\cite{jiang2018learning, yao2017boosting}. Qu et al.~\cite{qu2016deep} pioneer this direction by combining CNN and RNN/LSTM architectures while introducing two foundational datasets: UCM and Sydney. However, these early approaches struggled with the complexity of RSI and the need for fine-grained spatial understanding.
The second stage is marked by various attention mechanisms, starting with the attention-based approaches validated by Lu et al.~\cite{lu2019sound} that enable dynamic focus on different image regions. This leads to a series of specialized attention architectures: a multi-level attention model by Li et al.~\cite{li2020multi} addressing visual-semantic integration, attribute-guided and label-guided attention mechanisms by \cite{zhang2019description, zhang2019lam, liu2022remote}, and structured attention for precise pixel-level understanding by Zhao et al.~\cite{zhao2021high}. These advances significantly improved caption quality by better capturing the complex spatial relationships in RSI.
The third and current stage has seen the emergence of more sophisticated architectures focusing on comprehensive scene understanding. Notable advances include the explainable word-sentence framework~\cite{wang2020word}, two-stage multi-task learning approach~\cite{shen2020remote}, and semantic concept extraction method~\cite{li2024learning}. Recent work has particularly emphasized the integration of spatial information between different regions and the exploitation of contextual information in the decoder, addressing previous limitations in generating accurate descriptions.

To support these developments, several datasets have been created specifically for RSIC. The RSICD~\cite{lu2017exploring} and NWPU-Captions dataset stands as the largest collection with over 10,000 images and corresponding captions, while Sydney~\cite{qu2016deep} and UCM~\cite{qu2016deep} provide more focused collections of remote sensing scenes. However, a critical limitation of these datasets is their English-centric nature. Unlike natural image captioning, where multi-lingual datasets like STAIR Captions~\cite{yoshikawa2017stair} and COCO-CN~\cite{li2019coco} have enabled multi-lingual research, RSIC lacks parallel datasets in multiple languages. This absence of multi-lingual RSIC resources has significantly hindered the development of multi-lingual RSIC systems.

The challenges in RSIC extend beyond just dataset limitations. The field faces unique technical challenges including scale variance, orientation ambiguity, and the need for precise geographic and structural terminology. When considering multi-lingual scenarios, these challenges are further compounded by the need to accurately translate specialized terminology and maintain consistent spatial descriptions across different languages. This gap in multi-lingual RSIC research presents a significant opportunity for advancing the field toward more inclusive and globally applicable solutions.

\subsection{Large Vision-Language Models}
Recent advances in large vision-language models (LVLMs)~\cite{internvl2,wang2024qwen2,yao2024minicpm,internvl2.5,lu2024deepseek,llava} have revolutionized multi-modal understanding. These models, trained on massive datasets often containing billions of image-text pairs, demonstrate remarkable zero-shot capabilities and cross-modal understanding. The scale and diversity of training data have enabled these models to develop robust visual understanding and flexible language generation capabilities.
Prominent models include CLIP~\cite{radford2021learning}, which pioneered large-scale contrastive learning between images and text, establishing new benchmarks in visual-semantic alignment. GPT-4V~\cite{yang2023dawn} extends language model capabilities to visual inputs, demonstrating impressive performance across a wide range of visual tasks. LLaVA~\cite{llava} demonstrates strong performance in visual instruction following and image understanding tasks through efficient adaptation of large language models to visual inputs. Other notable models include Qwen2-VL~\cite{wang2024qwen2} and  InternVL2~\cite{internvl2} enhancing the visual perception of VLMs.
In image captioning, VLMs have shown impressive capabilities in generating detailed, contextually rich descriptions.
The emergence of LVLMs presents both opportunities and challenges for multi-lingual RSIC. While these models demonstrate potential in zero-shot inference, supervised fine-tuning, and cross-dataset transfer learning scenarios, their performance in specialized domains like RSI remains largely unexplored, particularly in multi-lingual settings. A systematic evaluation of state-of-the-art LVLMs on remote sensing tasks across different languages is crucial for understanding their capabilities and limitations in this specialized domain~\cite{hu2023rsgpt,kuckreja2024geochat,zhan2024skyeyegpt}.

\section{Construction of linguistic diversity}
\subsection{Datasets Annotation}

\begin{figure}[htb]
    \centering
    \includegraphics[width=0.48\textwidth]{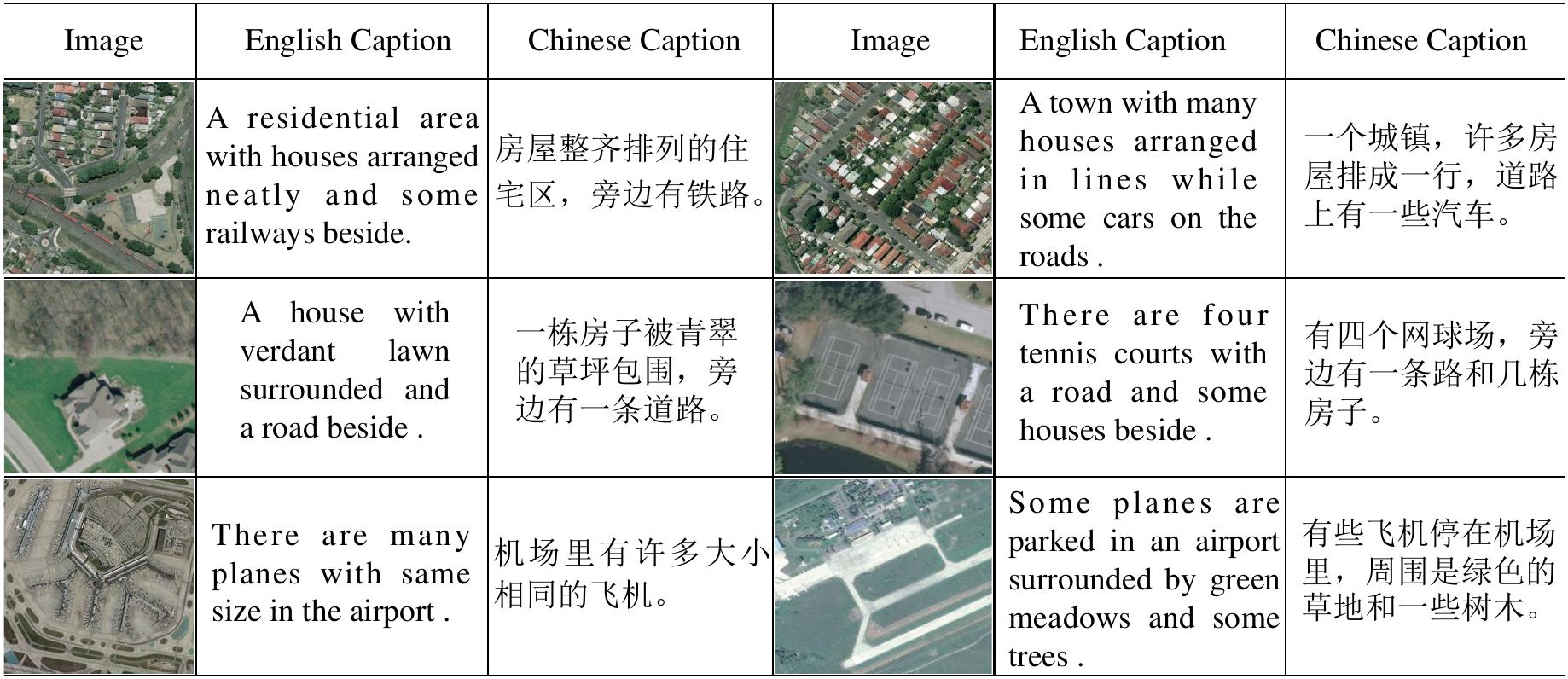}
    \caption{Sample of BRSIC, from top to bottom: Sydney, UCM, RSICD.}
    \label{fig:sample}
\end{figure}

\begin{figure*}[htb]
    \centering
    \includegraphics[width=0.9\textwidth]{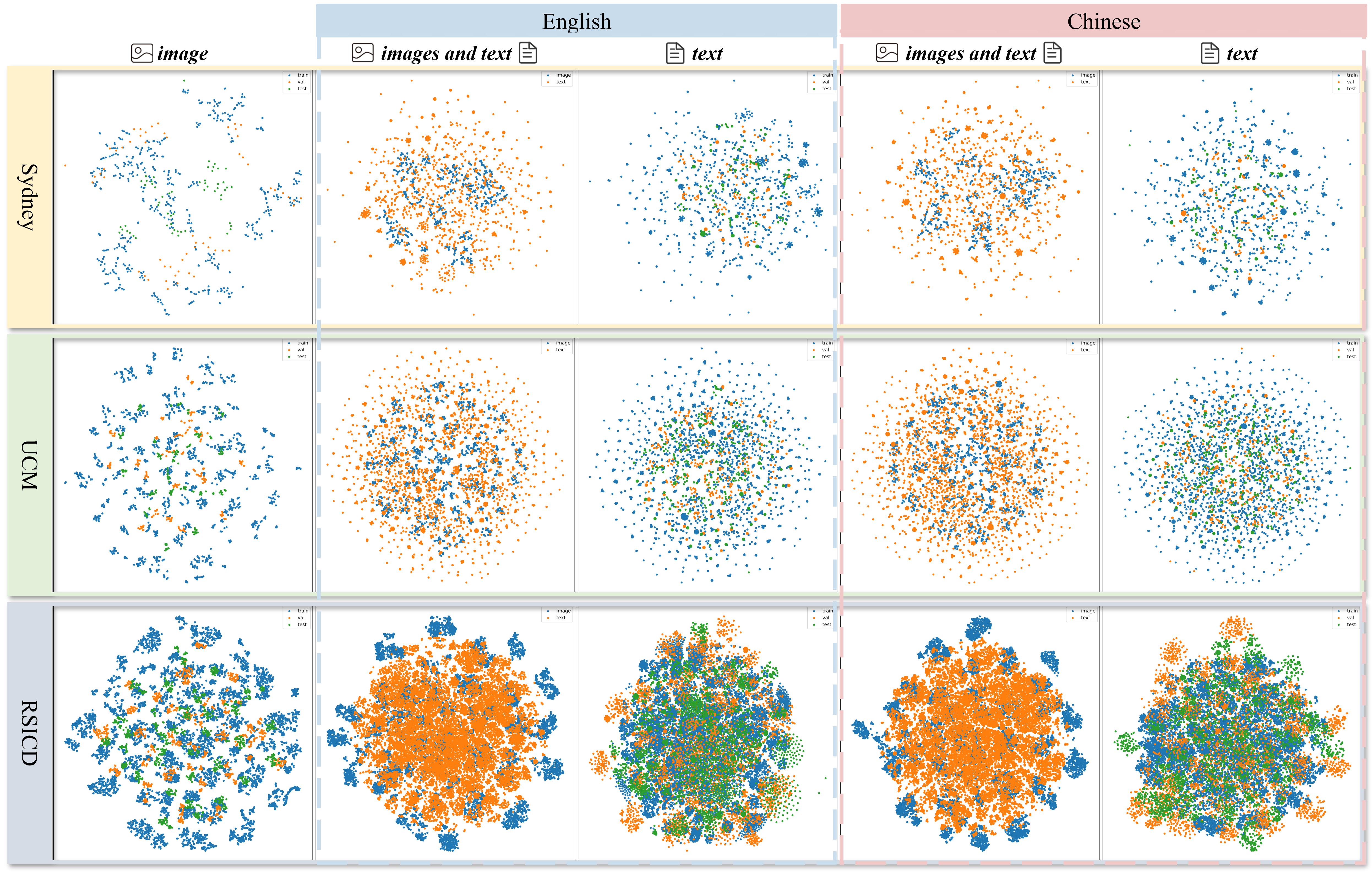}
    \caption{Visualization of textual and visual feature distribution of BRSIC based on t-SNE.}
    \label{fig:dataset_tsne}
\end{figure*}

The construction of BRSIC dataset involves annotating Chinese descriptions for three widely-used English RSIC datasets: UCM-Captions~\cite{qu2016deep}, Sydney-Captions~\cite{qu2016deep}, and RSICD~\cite{lu2017exploring}. The resulting bilingual dataset comprises 13,634 images paired with 68,170 captions in both English and Chinese, establishing a comprehensive resource for multi-lingual RSIC research.

\subsubsection{Dataset Overview}
The base datasets represent diverse remote sensing scenarios:
\begin{itemize}
    \item Sydney-Captions includes 613 high-resolution images (500×500 pixels) from Sydney, Australia, spanning 7 categories, accompanied by 3,065 English descriptions.
    \item UCM-Captions consists of 2,100 images (256×256 pixels) across 21 diverse scene categories, each paired with 5 English captions, totaling 10,500 captions.
    \item RSICD comprises 10,921 images (224×224 pixels) collected from various map services (Google Maps, Baidu Maps, etc.), encompassing 30 scene categories with 54,605 English captions.
\end{itemize}

\subsubsection{Annotation Pipeline}
Our annotation process employs a hybrid approach combining machine translation and manual correction to ensure both efficiency and quality. This methodology significantly reduces the annotation time and cost while maintaining high-quality standards.

We initially experimented with state-of-the-art machine translation systems (including Google Translate and Baidu Translate) to generate preliminary Chinese descriptions. However, we identified several critical limitations in direct machine translation:
\begin{itemize}
    \item Contextual misinterpretation (e.g., "planes park in the parking lot" being literally translated without considering the aviation context).
    \item Structural complexity in translating English subordinate clauses into Chinese.
\end{itemize}

Professional annotators with expertise in both remote sensing and bilingual translation review and correct the machine-translated descriptions. The correction process focus on accurate translation of domain-specific terminology, natural expression of spatial relationships in Chinese, and adaptation of sentence structures to match Chinese linguistic patterns.

\subsubsection{Annotation Characteristics}
The resulting Chinese annotations exhibit several distinctive features compared to their English counterparts. First, they employ simplified sentence structures that better align with Chinese linguistic conventions, particularly in describing spatial relationships. Second, they adapt complex English subordinate clauses into more straightforward Chinese expressions to improve readability. Third, they maintain consistent terminology usage while accommodating Chinese language patterns and conventions.

As shown in Fig.~\ref{fig:sample}, this comprehensive annotation process results in a high-quality bilingual dataset that preserves the semantic content of the original English descriptions while incorporating natural Chinese linguistic characteristics when appropriate.

\subsection{Datasets Analysis}

\subsubsection{Feature Distribution}
To analyze the feature relationships across modalities and languages in BRSIC, we employ t-SNE dimensionality reduction to visualize the distribution patterns of visual and textual features. The resulting visualization matrix (Fig. \ref{fig:dataset_tsne}) comprises feature projections (extracted by CLIP~\cite{radford2021learning}) for Sydney, UCM, and RSICD datasets, revealing three key characteristics:

The cross-modal feature relationships demonstrate distinct patterns across the visualization matrix. Visual features (leftmost column) exhibit more concentrated clustering patterns compared to textual features, suggesting strong intra-class visual similarities across all three datasets. In contrast, textual features, both in English and Chinese, show more dispersed distributions, reflecting the natural variance in language expressions for describing similar visual content.

The cross-lingual semantic preservation is evidenced by the remarkable similarity between English and Chinese feature distributions. Both languages maintain consistent global structures within each dataset, with corresponding clusters appearing in similar regions of the feature space. Minor variations in cluster shapes between languages reflect natural linguistic differences rather than semantic inconsistencies, validating our annotation pipeline's effectiveness.

Dataset-specific characteristics are clearly visible and correlate with both the inherent properties and scales of each dataset. Sydney dataset (613 images) exhibits scattered, well-separated clusters, characteristic of its diverse high-resolution urban scenes and relatively focused domain coverage. UCM (2,100 images) shows more structured and compact groupings, aligning with its carefully curated scene categories and moderate dataset size. RSICD (10,921 images) demonstrates the highest feature density and inter-cluster connectivity, reflecting both its comprehensive coverage of remote sensing scenarios and substantially larger scale. These distribution patterns suggest that dataset size significantly influences feature space complexity, with larger datasets showing more intricate inter-class relationships.

A notable observation in RSICD dataset reveals what we term the "Cross-split Distribution Misalignment" (CSDM), a phenomenon where significant distributional discrepancies exist among training, validation, and test sets. While the training set exhibits a relatively concentrated distribution, both validation and test sets display more dispersed patterns, with portions extending beyond the training distribution boundaries. This misalignment is particularly pronounced in the validation set, which contains a higher proportion of out-of-distribution samples compared to the test set, resulting in substantial performance variations between validation and test evaluations. Li et al.~\cite{li2020multi} correct annotation errors, along with swapping the validation and test sets. These modifications have led to significant performance differences between the original and modified test sets, with models typically achieving notably lower performance on the original test set (can be seen from the Table~\ref{tab:vlm-rsicd}). For consistency with recent literature and to benefit from the corrected annotations, our annotation pipeline and experiments are based on the modified version of RSICD by default.

\begin{figure*}
    \centering
    \begin{subfigure}[b]{0.32\textwidth}
        \centering
        \includegraphics[height=3.2cm]{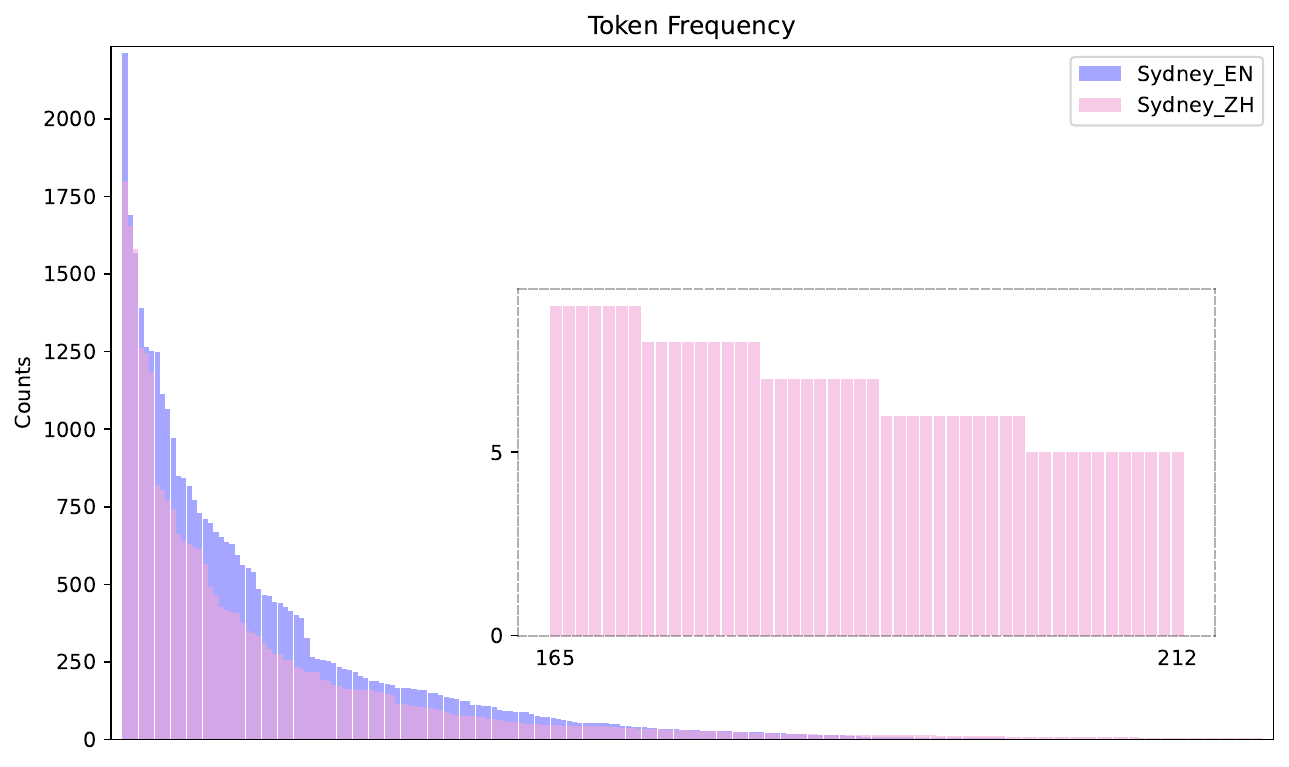}
        \caption{Sydney}
    \end{subfigure}
    \hfill
    \begin{subfigure}[b]{0.32\textwidth}
        \centering
        \includegraphics[height=3.2cm]{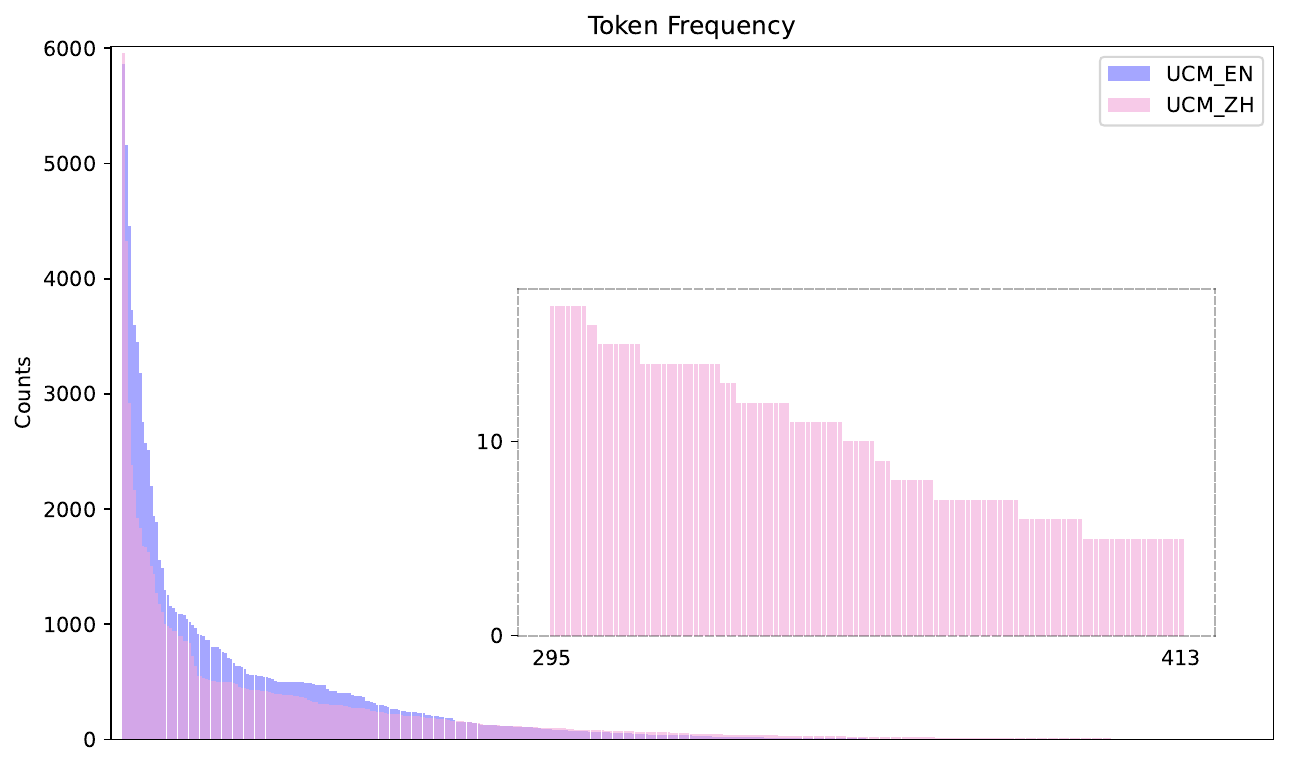}
        \caption{UCM}
    \end{subfigure}
    \hfill
    \begin{subfigure}[b]{0.32\textwidth}
        \centering
        \includegraphics[height=3.2cm]{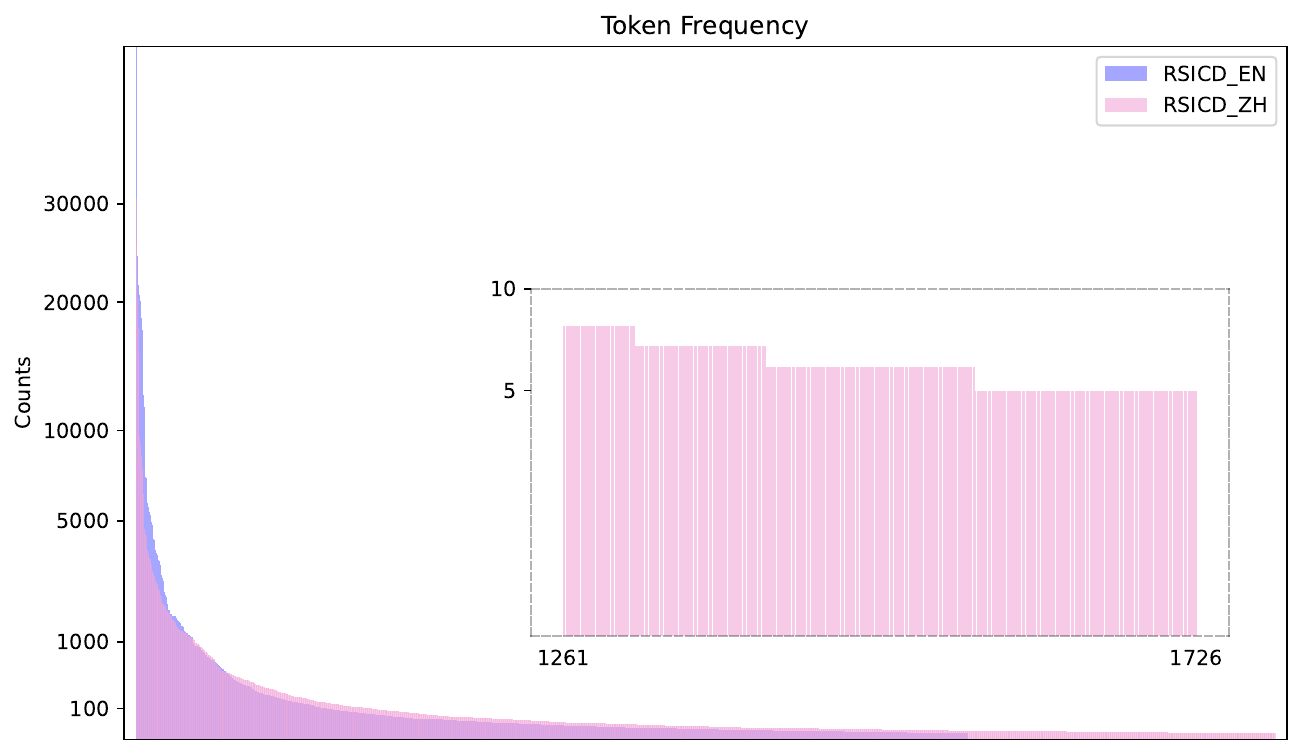}
        \caption{RSICD}
    \end{subfigure}
    \caption{Token distribution of BRSIC.}
    \label{fig:dataset_wordmap}
\end{figure*}

\begin{figure}
    \centering
    \includegraphics[width=0.45\textwidth]{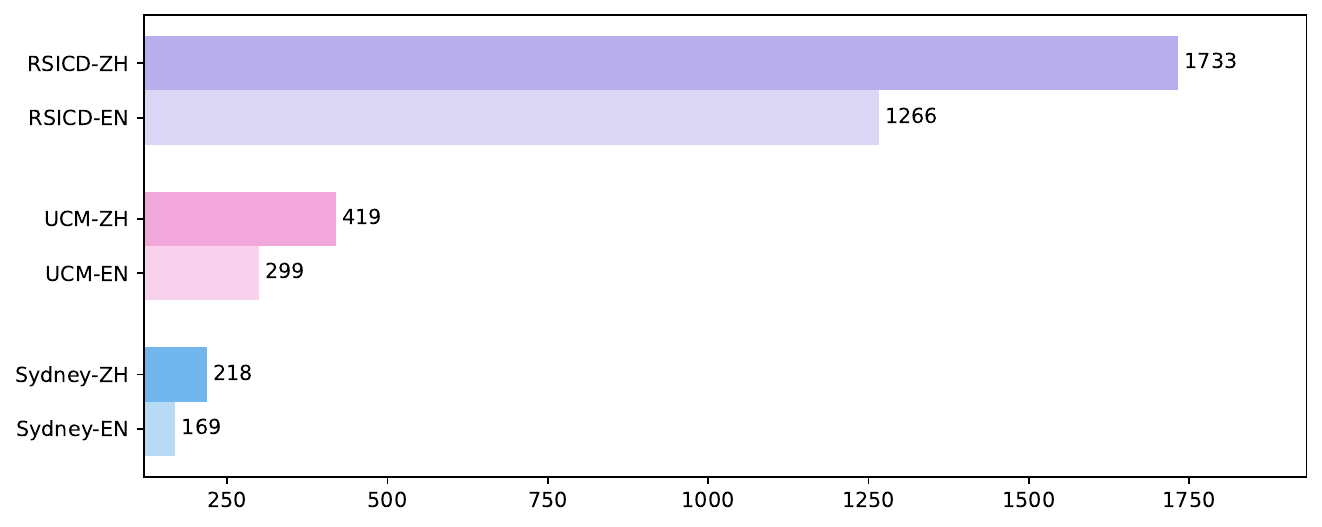}
    \caption{Vocabulary size of BRSIC.}
    \label{fig:dataset_wordmap_length}
\end{figure}

\begin{figure*}
    \centering
    \begin{subfigure}[b]{0.32\textwidth}
        \centering
        \includegraphics[width=0.9\textwidth]{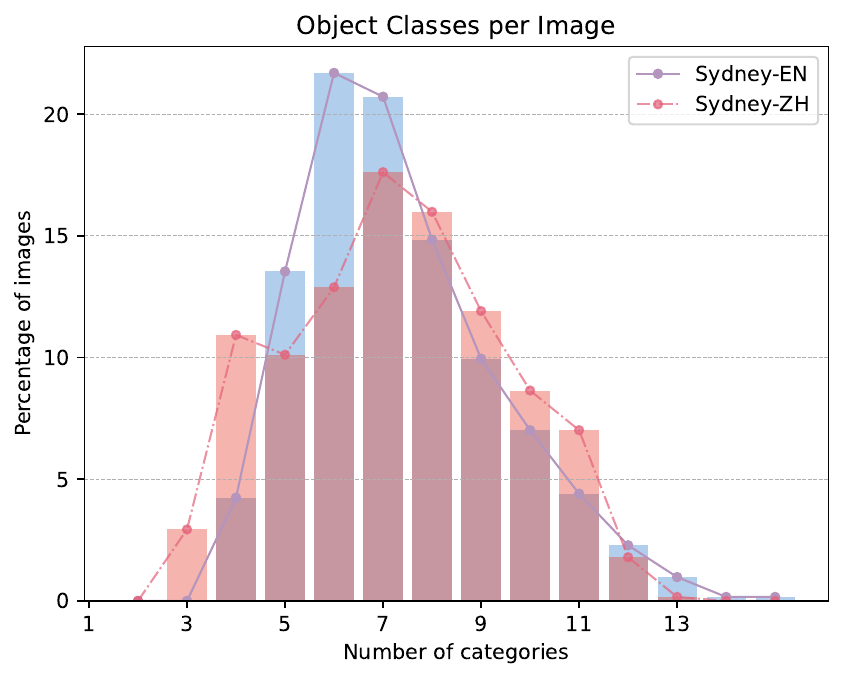}
        \caption{Sydney}
    \end{subfigure}
    \hfill
    \begin{subfigure}[b]{0.32\textwidth}
        \centering
        \includegraphics[width=0.9\textwidth]{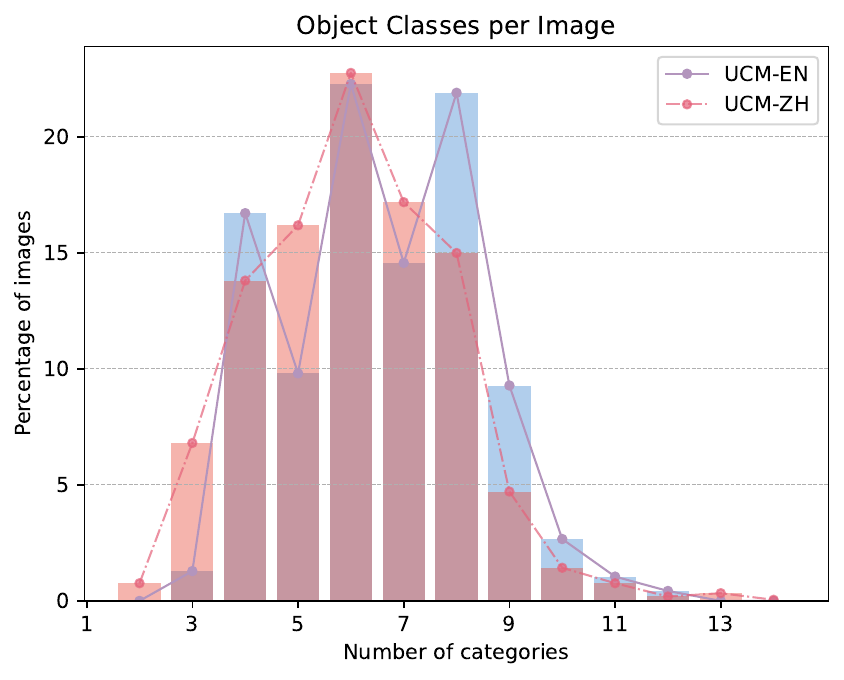}
        \caption{UCM}
    \end{subfigure}
    \hfill
    \begin{subfigure}[b]{0.32\textwidth}
        \centering
        \includegraphics[width=0.9\textwidth]{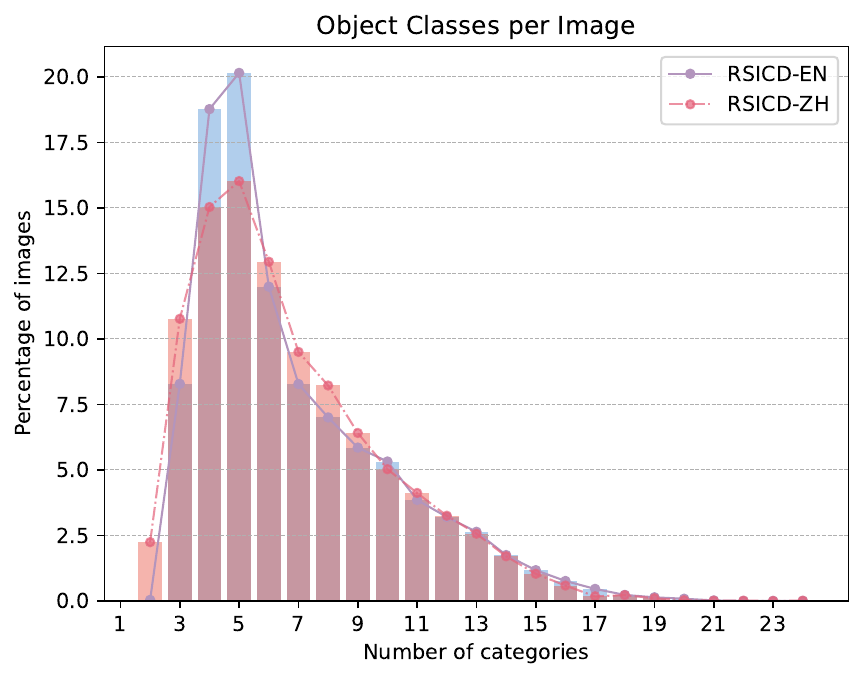}
        \caption{RSICD}
    \end{subfigure}
    \caption{Object number distribution of BRSIC.}
    \label{fig:dataset_object}
\end{figure*}

\begin{figure*}
    \centering
    \includegraphics[width=1\textwidth]{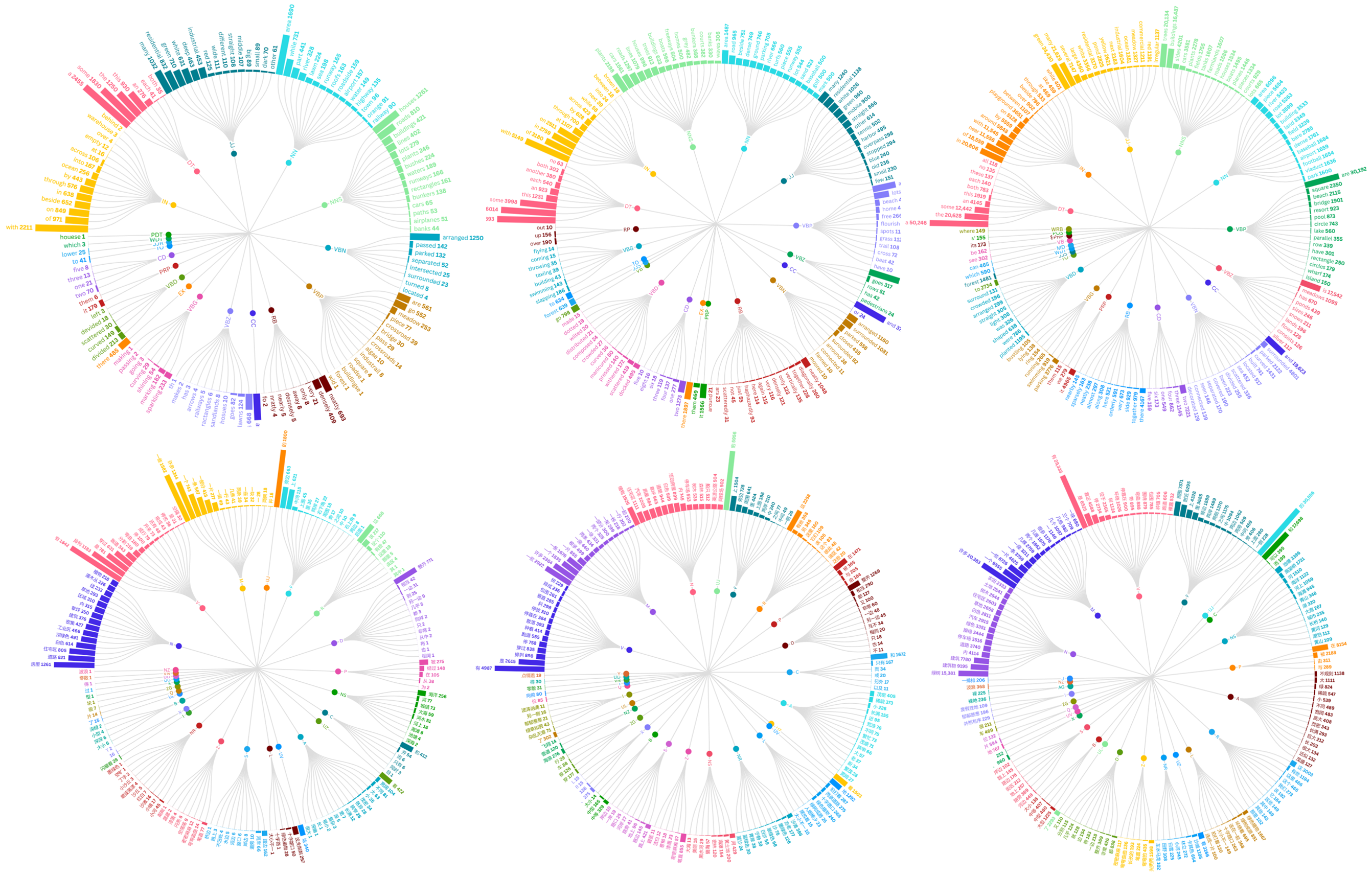}
    \caption{Part of Speech distribution of BRSIC (from left to right: Sydney, UCM, RSICD; from top to bottom: English, Chinese).}
    \label{fig:dataset_pos}
\end{figure*}

\subsubsection{Token Distribution}
Analysis of token distributions (Fig. \ref{fig:dataset_wordmap}) and vocabulary statistics (Fig. \ref{fig:dataset_wordmap_length}) reveals significant linguistic divergences between English and Chinese captions across the datasets.

English captions demonstrate concentrated token distributions with clear dominant terms, while Chinese descriptions exhibit broader token dispersion. This linguistic divergence is particularly evident in the RSICD dataset (Fig. \ref{fig:dataset_wordmap}c), which contains the largest vocabulary. A single English word often requires a combination of Chinese characters to express the same meaning, and these character combinations are tokenized as distinct units during text processing, naturally leading to both more dispersed token distributions and larger vocabulary sizes, as confirmed by Fig. \ref{fig:dataset_wordmap_length}.

These linguistic characteristics provide insights into the performance gap between Chinese and English in subsequent experiments. The increased vocabulary complexity and dispersed token distribution, both stemming from Chinese tokenization of character combinations, requires models to learn more sophisticated semantic relationships in the visual-textual mapping task.

\subsubsection{Object Number Distribution}

Analysis of object number distributions (Fig. \ref{fig:dataset_object}) reveals dataset-specific characteristics and subtle cross-lingual differences. The three datasets demonstrate distinct distribution patterns in terms of object numbers per image.

Sydney dataset shows a relatively broad distribution with its peak at 7-8 object categories per image, reflecting its high-resolution urban scenes that naturally contain more objects. UCM exhibits a more concentrated distribution peaking at 5-7 categories, aligning with its moderate scene complexity. RSICD demonstrates a right-skewed distribution with a peak at 4-5 categories, suggesting more focused scene descriptions despite its larger scale.

Cross-lingual comparison reveals subtle but consistent differences. Chinese annotations tend to show slightly higher frequencies at lower object numbers compared to their English counterparts, particularly visible in Sydney and UCM datasets. This systematic shift likely stems from two factors: differences in tokenization approaches, where multiple objects in English might be combined into a single compound token in Chinese, and inherent linguistic conventions where Chinese descriptions may favor more concise object enumeration. However, the overall distribution patterns remain largely parallel between language pairs, indicating the preservation of core semantic content during annotation.

\newcommand{\demph}[1]{\textcolor{gray}{#1}}

\begin{table*}
    \caption{Performance of traditional vision-language models on BRSIC-Sydney. English-Ori represents the results from the unmodified original English dataset.}
    \label{tab:vlm-sydney}
    \centering
    \setlength{\tabcolsep}{1.9pt}
    \begin{tabular}{l|ccccccc|ccccccc|ccccccc}
        \hline
        \multirow{2}{*}{Model} & \multicolumn{7}{c|}{English-Ori} & \multicolumn{7}{c|}{English} & \multicolumn{7}{c}{Chinese} \\
        ~ & B@1 & B@2 & B@3 & B@4 & M & R & C & B@1 & B@2 & B@3 & B@4 & M & R & C & B@1 & B@2 & B@3 & B@4 & M & R & C \\
        \hline
        BUTD~\cite{butd} & \textbf{79.26} & \textbf{70.88} & \textbf{63.63} & \textbf{57.12} & \textbf{38.85} & \textbf{71.14} & \textbf{246.60}
        & \textbf{79.07} & \textbf{70.65} & \textbf{63.38} & \textbf{56.70} & \textbf{39.68} & \textbf{72.20} & \textbf{253.50}
        & 74.33 & 64.64 & 55.60 & 46.77 & 36.43 & 61.88 & 148.03
        \\

        AoANet~\cite{aoanet} & 76.29 & 67.83 & 60.43 & 53.91 & 37.38 & 69.69 & 235.92
            & 77.93 & 69.33 & 62.30 & 55.93 & 38.30 & 70.30 & 247.29
            & \textbf{77.81} & \textbf{70.25} & \textbf{57.44} & \textbf{58.88} & \textbf{43.02} & \textbf{68.58} & \textbf{188.40}
        \\

        MLAT~\cite{mlat} & 73.41 & 65.63 & 59.12 & 53.23 & 37.21 & 67.50 & 218.85
            & 73.64 & 66.00 & 60.37 & 54.64 & 37.52 & 67.73 & 221.93
            & 68.17 & 56.82 & 49.21 & 43.82 & 32.02 & 61.48 & 167.59
        \\

        MG~\cite{MG} & 74.58 & 65.78 & 58.59 & 52.41 & 37.14 & 67.36 & 224.10
            & 76.86 & 68.55 & 61.34 & 54.86 & 38.12 & 69.50 & 233.46
            & 71.36 & 59.26 & 50.80 & 43.87 & 36.73 & 59.24 & 154.74
        \\

        \demph{MG*~\cite{MG}} & \demph{83.28} & \demph{77.30} & \demph{72.06} & \demph{67.53} & \demph{46.43} & \demph{77.99} & \demph{296.84}
            & \demph{84.51} & \demph{78.92} & \demph{73.77} & \demph{68.96} & \demph{44.84} & \demph{75.30} & \demph{260.49}
            & \demph{77.81} & \demph{70.25} & \demph{58.00} & \demph{51.54} & \demph{39.92} & \demph{63.97} & \demph{181.33}
        \\
        \hline
    \end{tabular}
\end{table*}

\begin{table*}
    \caption{Performance of traditional vision-language models on BRSIC-UCM.}
    \label{tab:vlm-ucm}
    \centering
    \setlength{\tabcolsep}{1.9pt}
    \begin{tabular}{l|ccccccc|ccccccc|ccccccc}
        \hline
        \multirow{2}{*}{Model} & \multicolumn{7}{c|}{English-Ori} & \multicolumn{7}{c|}{English} & \multicolumn{7}{c}{Chinese} \\
        ~ & B@1 & B@2 & B@3 & B@4 & M & R & C & B@1 & B@2 & B@3 & B@4 & M & R & C & B@1 & B@2 & B@3 & B@4 & M & R & C \\
        \hline
        BUTD~\cite{butd} & 80.55 & 74.32 & 69.26 & 64.83 & 42.15 & 76.70 & 312.37
            & 80.56 & 73.97 & 68.62 & 64.03 & 41.94 & 76.22 & 307.82
            & 70.41 & 62.10 & 55.80 & 50.30 & 37.01 & 65.35 & 192.69
        \\

        AoANet~\cite{aoanet} & 84.82 & 78.73 & 73.29 & 67.80 & 43.30 & 78.48 & \textbf{357.76}
            & 85.10 & 79.50 & 74.18 & 68.84 & 46.41 & \textbf{82.77} & \textbf{373.16}
            & 74.11 & 66.15 & 60.43 & 55.45 & \textbf{38.90} & 65.96 & 223.03
        \\
        MLAT~\cite{mlat} & \textbf{86.63} & 80.64 & 75.34 & 70.31 & 45.63 & 81.26 & 329.95
            & \textbf{86.99} & \textbf{81.99} & \textbf{77.44} & \textbf{73.17} & \textbf{47.04} & 81.59 & 342.02
            & \textbf{77.41} & \textbf{69.31} & \textbf{62.77} & \textbf{57.39} & 38.31 & \textbf{72.32} & \textbf{230.48}
        \\

        MG~\cite{MG} & 86.50 & \textbf{81.39} & \textbf{76.70} & \textbf{72.41} & \textbf{47.67} & \textbf{82.10} & 356.22
            & 85.15 & 79.74 & 74.88 & 70.43 & 45.23 & 80.23 & 351.77
            & 70.64 & 62.79 & 56.40 & 50.96 & 39.90 & 65.81 & 213.96
        \\

        \demph{MG*~\cite{MG}} & \demph{89.04} & \demph{84.81} & \demph{81.04} & \demph{77.71} & \demph{51.97} & \demph{85.91} & \demph{389.88}
            & \demph{89.80} & \demph{86.03} & \demph{82.67} & \demph{79.68} & \demph{53.39} & \demph{86.59} & \demph{410.39}
            & \demph{81.30} & \demph{76.19} & \demph{71.92} & \demph{68.14} & \demph{47.09} & \demph{74.50} & \demph{270.11}
        \\
        \hline
    \end{tabular}
    \end{table*}

\begin{table*}
    \caption{Performance of traditional vision-language models on BRSIC-RSICD.}
    \label{tab:vlm-rsicd}
    \centering
    \setlength{\tabcolsep}{1.9pt}
    \begin{tabular}{l|ccccccc|ccccccc|ccccccc}
        \hline
        \multirow{2}{*}{Model} & \multicolumn{7}{c|}{English-Ori} & \multicolumn{7}{c|}{English} & \multicolumn{7}{c}{Chinese} \\
        ~ & B@1 & B@2 & B@3 & B@4 & M & R & C & B@1 & B@2 & B@3 & B@4 & M & R & C & B@1 & B@2 & B@3 & B@4 & M & R & C \\
        \hline
        BUTD~\cite{butd} & 62.13 & 46.42 & 36.81 & 30.22 & 25.36 & 48.18 & 80.16
            & 75.85 & 64.01 & 54.63 & 47.04 & 35.79 & 65.48 & 256.36
            & 61.08 & 48.87 & 39.48 & 32.37 & 30.04 & 49.50 & 150.49
        \\

        AoANet~\cite{aoanet} & 62.85 & 46.26 & 35.86 & 28.92 & 25.15 & 47.03 & 80.84
            & 69.97 & 57.38 & 47.86 & 40.36 & 34.51 & 61.71 & 221.08
            & 58.13 & 44.79 & 35.15 & 27.94 & 29.53 & 46.82 & 131.60
        \\
        MLAT~\cite{mlat} & 65.56 & \textbf{49.53} & \textbf{39.46} & \textbf{32.54} & 26.64 & \textbf{49.81} & \textbf{90.92}
            & 75.72 & 65.55 & 57.26 & 50.20 & 38.25 & 67.63 & 249.30
            & 61.15 & 46.97 & 37.92 & 31.12 & 26.45 & 50.73 & 153.42
        \\

        MG~\cite{MG} & \textbf{65.40} & 48.82 & 38.46 & 31.86 & \textbf{26.93} & 49.21 & 84.68
            & \textbf{79.29} & \textbf{68.46} & \textbf{52.92} & \textbf{52.23} & \textbf{38.88} & \textbf{69.45} & \textbf{286.06}
            & \textbf{63.09} & \textbf{51.87} & \textbf{43.02} & \textbf{36.16} & \textbf{31.50} & \textbf{51.90} & \textbf{168.33}
        \\

        \demph{MG*~\cite{MG}} & \demph{70.97} & \demph{60.35} & \demph{52.44} & \demph{46.33} & \demph{34.82} & \demph{62.89} & \demph{288.51}
            & \demph{69.78} & \demph{58.69} & \demph{50.44} & \demph{44.10} & \demph{33.67} & \demph{61.73} & \demph{277.98}
            & \demph{55.19} & \demph{44.99} & \demph{37.49} & \demph{31.82} & \demph{28.71} & \demph{49.11} & \demph{166.10}
        \\
        \hline
    \end{tabular}
    \end{table*}

\subsubsection{Part of Speech Distribution}

Analysis of Part-of-Speech (POS) distributions (Fig. \ref{fig:dataset_pos}) reveals both dataset characteristics and multi-lingual patterns. The circular visualizations demonstrate systematic differences in syntactic structures between English and Chinese captions across all three datasets.

In English captions (top row), nouns and prepositions dominate the distributions, reflecting the spatial-descriptive nature of remote sensing image captions. The prominence of prepositions (e.g., "in", "on", "beside") indicates English's reliance on explicit spatial relationship markers. In contrast, Chinese captions (bottom row) show a higher proportion of verbs and adjectives relative to prepositions, as spatial relationships in Chinese are often encoded through verb-based constructions rather than prepositional phrases.

Dataset-wise comparisons reveal consistent patterns. Sydney and UCM datasets exhibit similar POS distributions due to their focused urban scene, while RSICD shows more diverse POS patterns, reflecting its broader coverage of remote sensing scenarios. This consistency across datasets suggests that the observed cross-lingual differences stem from fundamental linguistic structures rather than dataset-specific characteristics.

These syntactic differences provide additional context for understanding multi-lingual model performance, as they indicate distinct ways of encoding spatial relationships and scene descriptions between English and Chinese.
\section{Assessment of multi-lingual adaptability}

\subsection{Evaluation Setup}
This study establishes a comprehensive evaluation framework to assess the multi-lingual adaptability of vision-language models in remote sensing image captioning. The evaluation metrics include BLEU@1-4~\cite{BLEU}, METEOR~\cite{METEOR}, ROUGE-L~\cite{ROUGE_L}, and CIDEr~\cite{CIDEr}, which collectively measure both the fluency and semantic accuracy of generated captions.
The experimental involves re-training and testing four traditional VLMs~\cite{butd,aoanet,MG,mlat} on the original English version, modified version, and Chinese version to demonstrate the impact of the modified version of the dataset on model performance, establishing a fair comparison.
The evaluation also includes eight state-of-the-art LVLMs~\cite{internvl2,wang2024qwen2,llava,lu2024deepseek}, ranging from 1.3B to 8B parameters, under three distinct scenarios: zero-shot inference, supervi sed fine-tuning, and multi-lingual training. We also compare the performance of VLMs and LVLMs under cross-dataset transfer scenarios.
All traditional VLMs are trained using their default hyperparameters. For LVLMs, we employ the AdamW optimizer with a learning rate of 1e-4, a batch size of 16 for $\sim$1B models, and 8 for $\sim$7B models, alongside a weight decay of 0.1. To optimize GPU memory usage, we incorporate LoRA\cite{hu2022lora} with a rank of 8 and utilize the ZeRO-2 stage for distributed training.

\subsection{Traditional Vision-Language Models}
As shown in Table~\ref{tab:vlm-sydney},~\ref{tab:vlm-ucm}, and~\ref{tab:vlm-rsicd}, we evaluate four representative traditional vision-language models: Bottom-Up-Top-Down (BUTD)~\cite{butd}, Attention on Attention Network (AoANet)~\cite{aoanet}, Multilayer Aggregated Transformer (MLAT)~\cite{mlat}, and Multiscale Grouping Transformer (MG)~\cite{MG}. These models are trained and tested on three variants of each dataset: the original English version, modified English version, and the Chinese version. The results reveal distinct patterns in cross-lingual performance.

\begin{table*}
    \caption{Performance of large vision-language models on BRSIC-Sydney.}
    \label{tab:sydney}
    \centering
    \setlength{\tabcolsep}{4pt}
    \begin{tabular}{l|ccccccc|ccccccccc}
        \toprule
        \multirow{2}{*}{Model} & \multicolumn{7}{c|}{English} & \multicolumn{7}{c}{Chinese} \\
        ~ & B@1 & B@2 & B@3 & B@4 & M & R & C & B@1 & B@2 & B@3 & B@4 & M & R & C \\
        \midrule
        \multicolumn{15}{c}{\textit{Zero-shot}} \\
        \midrule
        InternVL2-1B~\cite{internvl2} & 25.05 & 13.55 & 9.15 & 6.34 & 15.28 & 29.39 & 8.53 
            & 0.00 & 0.00 & 0.00 & 0.00 & 0.00 & 0.00 & 0.00 \\
        InternVL2-2B~\cite{internvl2} & 39.47 & 16.98 & 8.76 & 5.09 & 13.24 & 32.18 & 15.48 
            & \textbf{41.24} & 18.35 & 9.06 & 4.68 & 17.77 & 33.60 & 10.11 \\
        InternVL2-8B~\cite{internvl2} & 36.34 & 14.76 & 6.04 & 2.43 & 14.14 & 31.56 & 13.66 
            & 39.50 & 18.09 & 8.36 & 3.06 & \textbf{17.92} & 34.05 & 9.21 \\
        DeepseekVL-1.3B~\cite{lu2024deepseek} & 37.71 & \textbf{22.33} & \textbf{15.20} & \textbf{10.42} & 15.13 & 34.84 & \textbf{19.37} 
            & 0.00 & 0.00 & 0.00 & 0.00 & 9.98 & 7.66 & 0.00 \\
        DeepseekVL-7B~\cite{lu2024deepseek} & 33.01 & 11.45 & 6.04 & 3.78 & 11.33 & 27.34 & 18.56 
            & 14.48 & 2.22 & 0.00 & 0.00 & 9.96 & 19.55 & 5.82 \\
        LLaVA-1.5-7B~\cite{llava} & \textbf{39.97} & 21.78 & 13.93 & 8.71 & \textbf{15.31} & \textbf{35.04} & 14.90 
            & 15.45 & 4.78 & 1.36 & 0.61 & 14.80 & 30.11 & 10.15 \\
        Qwen2-VL-2B~\cite{wang2024qwen2} & 24.93 & 9.35 & 4.42 & 2.50 & 6.82 & 18.27 & 8.40 
            & 30.35 & 13.31 & 5.99 & 3.17 & 15.18 & 24.33 & 12.38 \\
        Qwen2-VL-7B~\cite{wang2024qwen2} & 37.70 & 16.36 & 8.40 & 4.77 & 14.68 & 32.94 & \textbf{19.37} 
            & 40.48 & \textbf{20.36} & \textbf{11.83} & \textbf{5.57} & \textbf{17.92} & \textbf{34.91} & \textbf{15.95} \\
        \midrule
        \multicolumn{15}{c}{\textit{Supervised Fine-tuning}} \\
        \midrule
        InternVL2-1B~\cite{internvl2} & 76.37 & 67.82 & 60.73 & 54.39 & 38.15 & 69.69 & 245.10 
            & 74.74 & 55.92 & 45.19 & 36.84 & 36.09 & 62.50 & 145.99 \\
        InternVL2-2B~\cite{internvl2} & 76.71 & 68.46 & 61.16 & 54.63 & \textbf{39.66} & 71.36 & 254.02 
            & 75.91 & 61.07 & 56.08 & 42.41 & 38.58 & 66.87 & 162.57 \\
        InternVL2-8B~\cite{internvl2} & 76.50 & 68.84 & 62.55 & 57.23 & 39.39 & 70.67 & 257.83 
            & 76.39 & 66.83 & 59.99 & 55.39 & 41.51 & 72.46 & 230.48  \\
        DeepseekVL-1.3B~\cite{lu2024deepseek} & 76.60 & 67.36 & 59.54 & 52.64 & 37.87 & 70.25 & 233.69 
            & \textbf{78.99} & 62.46 & 52.67 & 45.26 & 39.40 & 68.36 & 176.59 \\
        DeepseekVL-7B~\cite{lu2024deepseek} & 76.27 & 68.68 & 62.30 & 56.80 & 39.12 & \textbf{71.73} & \textbf{260.90} 
            & 74.65 & 62.94 & 54.22 & 48.30 & 39.82 & 68.75 & 211.00 \\
        LLaVA-1.5-7B~\cite{llava} & 77.72 & \textbf{70.02} & \textbf{63.82} & \textbf{58.49} & 39.33 & 71.02 & 249.31 
            & 73.78 & 63.35 & 56.10 & 51.02 & 38.91 & 69.15 & 200.73  \\
        Qwen2-VL-2B~\cite{wang2024qwen2} & 76.81 & 67.80 & 60.82 & 54.80 & 38.96 & 70.28 & 246.82 
            & 73.67 & 61.79 & 53.73 & 47.89 & 38.98 & 68.68 & 189.26 \\
        Qwen2-VL-7B~\cite{wang2024qwen2} & \textbf{78.14} & 69.40 & 61.98 & 55.32 & 39.32 & 70.96 & 249.03 
            & 77.12 & \textbf{67.94} & \textbf{60.99} & \textbf{56.33} & \textbf{42.89} & \textbf{74.11} & \textbf{251.52} \\
        \midrule

        \multicolumn{15}{c}{\textit{Multi-lingual Training}} \\
        \midrule
        InternVL2-1B~\cite{internvl2} & 76.54 & 68.36 & 61.52 & 55.50 & 39.25 & 70.64 & 245.37 
            & 74.00 & 58.83 & 49.21 & 42.26 & 38.30 & 66.40 & 179.78\\

        InternVL2-2B~\cite{internvl2} & 76.94 & 68.70 & 61.64 & 55.20 & 40.68 & 72.72 & \textbf{264.88} 
            & 77.16 & \textbf{65.43} & \textbf{57.09} & 50.39 & 40.71 & 70.59 & 205.20 \\
        InternVL2-8B~\cite{internvl2} & \textbf{79.89} & \textbf{72.07} & \textbf{65.10} & \textbf{58.82} & \textbf{41.35} & \textbf{72.83} & 252.14 
            & 77.02 & 64.30 & 56.95 & \textbf{51.31} & \textbf{41.97} & 71.25 & 204.16 \\

        DeepseekVL-1.3B~\cite{lu2024deepseek} & 79.76 & 71.52 & 64.38 & 57.73 & 40.07 & 73.01 & 263.21 
            & \textbf{80.79} & 65.33 & 55.72 & 48.12 & 41.11 & 70.61 & 189.14 \\

        DeepseekVL-7B~\cite{lu2024deepseek} & 77.48 & 69.80 & 63.02 & 56.73 & 40.38 & 71.33 & 253.30 
            & 74.29 & 62.46 & 55.22 & 50.07 & 40.62 & 69.95 & 199.18 \\
        LLaVA-1.5-7B~\cite{llava} & 78.14 & 68.94 & 61.13 & 53.99 & 38.03 & 70.61 & 238.35 
            & 75.45 & 59.96 & 51.57 & 45.26 & 39.50 & 67.94 & 171.12 \\

        Qwen2-VL-2B~\cite{wang2024qwen2} & 77.41 & 69.29 & 62.09 & 55.65 & 39.39 & 72.05 & 251.94 
            & 75.21 & 64.35 & 56.75 & 50.95 & 41.17 & 70.96 & \textbf{217.02} \\
        Qwen2-VL-7B~\cite{wang2024qwen2} & 76.67 & 67.76 & 60.22 & 52.74 & 38.86 & 73.01 & 225.97 
            & 77.78 & 63.45 & 55.01 & 48.12 & 41.37 & \textbf{71.73} & 208.93 \\
        \bottomrule

    \end{tabular}
    \end{table*}

\begin{table*}
    \caption{Performance of large vision-language models on BRSIC-UCM.}
    \label{tab:ucm}
    \centering
    \setlength{\tabcolsep}{4pt}
    \begin{tabular}{l|ccccccc|ccccccccc}
    \toprule
        \multirow{2}{*}{Model} & \multicolumn{7}{c|}{English} & \multicolumn{7}{c}{Chinese} \\
        ~ & B@1 & B@2 & B@3 & B@4 & M & R & C & B@1 & B@2 & B@3 & B@4 & M & R & C \\
    \midrule
        \multicolumn{15}{c}{\textit{Zero-shot}} \\
        \midrule
        InternVL2-1B~\cite{internvl2} & 23.32 & 11.34 & 6.58 & 3.73 & 14.58 & 24.60 & 13.61 
            & 0.00 & 0.00 & 0.00 & 0.00 & 0.00 & 0.00 & 0.00 \\
        InternVL2-2B~\cite{internvl2} & 34.83 & 17.71 & 11.12 & 6.49 & 12.29 & 28.85 & 30.47 
            & 40.04 & 19.35 & 10.19 & 4.01 & 18.04 & 34.86 & 18.72 \\
        InternVL2-8B~\cite{internvl2} & 33.83 & 16.87 & 9.36 & 4.99 & 13.65 & 29.92 & 30.15 
            & 40.12 & 19.37 & 10.47 & 5.47 & 17.02 & 35.27 & 16.38 \\
        DeepseekVL-1.3B~\cite{lu2024deepseek} & 34.48 & 17.55 & 10.18 & 5.90 & 14.18 & 29.03 & 33.74 
            & 0.35 & 0.18 & 0.08 & 0.00 & 1.16 & 3.94 & 6.36 \\
        DeepseekVL-7B~\cite{lu2024deepseek} & 32.87 & 15.50 & 8.96 & 5.09 & 13.82 & 27.26 & 34.63 
            & 23.34 & 10.63 & 4.97 & 2.48 & 13.66 & 21.70 & 19.59 \\
        LLaVA-1.5-7B~\cite{llava} & \textbf{37.61} & \textbf{20.25} & 11.24 & 5.55 & \textbf{16.11} & 31.93 & 39.57 
            & 38.61 & 0.00 & 0.00 & 0.00 & 18.58 & 42.21 & 0.24 \\
        Qwen2-VL-2B~\cite{wang2024qwen2} & 35.84 & 18.97 & 11.49 & 6.58 & 14.50 & 26.82 & 38.39 
            & 34.96 & 18.94 & 7.93 & 7.34 & 17.13 & 27.91 & 31.71 \\
        Qwen2-VL-7B~\cite{wang2024qwen2} & 36.65 & 19.76 & \textbf{12.07} & \textbf{7.53} & 15.90 & \textbf{32.71} & \textbf{45.66} 
            & \textbf{45.85} & \textbf{26.83} & \textbf{16.60} & \textbf{10.33} & \textbf{21.16} & \textbf{40.30} & \textbf{44.97} \\
    \midrule
        \multicolumn{15}{c}{\textit{Supervised Fine-tuning}} \\
        \midrule
        InternVL2-1B~\cite{internvl2} & 83.12 & 75.94 & 70.258 & 65.50 & 43.35 & 79.08 & 334.47 
            & 79.24 & 69.87 & 62.84 & 57.12 & 43.70 & 75.16 & 240.20  \\
        InternVL2-2B~\cite{internvl2} & 84.37 & 77.36 & 71.63 & 66.22 & 45.09 & 80.79 & 334.72 
            & 77.99 & 68.33 & 61.05 & 55.26 & 43.98 & 75.00 & 237.15 \\
        InternVL2-8B~\cite{internvl2} & 86.00 & 78.97 & 73.09 & 67.73 & 45.44 & 81.33 & \textbf{360.54} 
            & 79.75 & 70.64 & 64.06 & 58.81 & 44.78 & 74.94 & 241.75 \\
        DeepseekVL-1.3B~\cite{lu2024deepseek} & 87.62 & 80.94 & 74.87 & 69.02 & 46.27 & 83.05 & 355.99 
            & 80.24 & \textbf{72.11} & \textbf{65.90} & \textbf{60.63} & \textbf{47.75} & \textbf{79.05} & \textbf{259.86} \\
        DeepseekVL-7B~\cite{lu2024deepseek} & 87.89 & \textbf{81.80} & \textbf{76.65} & \textbf{72.06} & \textbf{47.29} & \textbf{83.39} & 356.42 
            & 79.77 & 71.54 & 65.30 & 60.13 & 45.55 & 76.12 & 237.98  \\
        LLaVA-1.5-7B~\cite{llava} & \textbf{88.22} & 81.61 & 75.60 & 69.93 & 44.83 & 81.95 & 336.54 
            & 78.45 & 70.03 & 63.64 & 58.44 & 44.34 & 76.50 & 221.79  \\
        Qwen2-VL-2B~\cite{wang2024qwen2} & 87.63 & 79.67 & 74.28 & 69.33 & 44.76 & 81.04 & 348.10 
            & \textbf{80.67} & 71.41 & 64.56 & 58.88 & 45.91 & 78.13 & 246.09 \\
        Qwen2-VL-7B~\cite{wang2024qwen2} & 87.06 & 80.43 & 74.71 & 69.42 & 45.38 & 82.33 & 347.15 
            & 80.07 & 70.99 & 63.93 & 57.91 & 45.93 & 77.45 & 257.42 \\
    
        \midrule
        \multicolumn{15}{c}{\textit{Multi-lingual Training}} \\
        \midrule
        InternVL2-1B~\cite{internvl2} & 84.48 & 76.49 & 70.02 & 64.24 & 43.86 & 78.92 & 342.87 
            & 79.69 & 70.33 & 63.57 & 58.11 & 44.26 & 74.45 & 242.59 \\
        InternVL2-2B~\cite{internvl2} & 83.43 & 75.92 & 69.77 & 64.31 & 44.59 & 78.99 & 346.65 
            & 79.34 & 70.69 & 64.23 & 59.09 & 45.15 & 75.96 & 242.60 \\
        InternVL2-8B~\cite{internvl2} & 86.73 & 80.17 & 75.34 & 70.96 & 45.83 & 80.74 & 362.61 
            & 81.31 & 73.05 & 66.46 & 60.87 & 47.12 & 78.29 & 258.15\\
        DeepseekVL-1.3B~\cite{lu2024deepseek} & 86.34 & 79.83 & 74.30 & 69.02 & 47.62 & 83.48 & 361.55 
            & 82.23 & 73.63 & 67.10 & 61.66 & 47.82 & 79.32 & 259.28 \\
        DeepseekVL-7B~\cite{lu2024deepseek} & 84.41 & 75.83 & 68.94 & 63.02 & 41.85 & 77.38 & 327.26 
            & 82.08 & 72.47 & 65.44 & 58.81 & 45.63 & 77.66 & 258.28 \\
        LLaVA-1.5-7B~\cite{llava} & \textbf{88.18} & 81.21 & 75.73 & 70.79 & 47.47 & 82.90 & 362.67 
            & 80.71 & 72.00 & 65.41 & 59.82 & 46.39 & 77.76 & 259.93 \\
        Qwen2-VL-2B~\cite{wang2024qwen2} & 84.64 & 77.52 & 71.04 & 65.16 & 44.01 & 80.18 & 329.66 
            & 77.62 & 68.76 & 62.56 & 57.78 & 45.92 & 76.19 & 248.99 \\
        Qwen2-VL-7B~\cite{wang2024qwen2} & 87.36 & \textbf{82.09} & \textbf{77.22} & \textbf{72.85} & \textbf{47.69} & \textbf{83.55} & \textbf{376.71} 
            & \textbf{84.77} & \textbf{76.78} & \textbf{70.91} & \textbf{66.06} & \textbf{47.84} & \textbf{79.88} & \textbf{277.63} \\
    \bottomrule
    \end{tabular}
    \end{table*}

\subsubsection{Comparison Across Methods} Among the four models, each demonstrates distinct strengths. BUTD achieves superior performance on the Sydney dataset, particularly in English caption generation (BLEU@1: 79.26, CIDEr: 246.60). MLAT exhibits exceptional performance on the UCM dataset, achieving the highest scores across multiple metrics (BLEU@1: 86.99, METEOR: 47.04). AoANet shows remarkable stability across different languages, while MG demonstrates strong adaptability on RSICD, achieving the best performance in both English (BLEU@1: 79.29, CIDEr: 286.06) and Chinese (BLEU@1: 63.09, CIDEr: 168.33) versions.

\subsubsection{Cross-version Analysis} The comparison between English-Ori and English versions reveals interesting patterns. On the Sydney and UCM datasets, the performance differences between these versions are minimal (e.g., BUTD: 79.26 vs 79.07 in BLEU@1), suggesting that dataset modification maintains the semantic integrity of the captions. However, on the RSICD dataset, we observe substantial performance improvements in the modified English version compared to the original (e.g., BUTD: 62.13 vs 75.85 in BLEU@1).
Due to the different test sets used by different methods, the performance differences are significant, which makes the comparison meaningless. We recommend that future studies specify which test set is used when comparing methods.
As shown in the t-SNE analysis (Figure~\ref{fig:dataset_tsne}), this significant gap can be attributed to the CSDM, where the modified test set exhibits closer distribution to the training set compared to the original version.

\subsubsection{Analysis of Random Split} To further validate our hypothesis about CSDM, we conducted additional experiments using random splitting strategy (denoted as MG*). The results show that under random splitting, the performance gap between English-Ori and English versions significantly narrows across all datasets (e.g., on RSICD, MG*: 70.97 vs 69.78 in BLEU@1). Meanwhile, on the modified version of RSICD, MG* achieves comparable performance to MG. These findings further validate our hypothesis about the impact of data distribution on model performance.

Specifically, while MG* demonstrates higher absolute scores in some cases (e.g., on Sydney, achieving BLEU@1 of 83.28 compared to MG's 74.58), the key insight lies in the consistency between versions under random splitting. This consistent performance across English-Ori and English versions, particularly evident in RSICD where the previous large gaps diminish, strongly suggests that the performance differences observed in our main experiments are indeed attributable to data distribution characteristics rather than inherent differences in caption quality. The random splitting results thus provide crucial empirical support for our analysis of CSDM's impact on model evaluation.

\subsection{Large Vision-Language Models}

\begin{table*}
    \caption{Performance of large vision-language models on BRSIC-RSICD.}
    \label{tab:rsicd}
    \centering
    \setlength{\tabcolsep}{4pt}
    \begin{tabular}{l|ccccccc|ccccccccc}
        \toprule
            \multirow{2}{*}{Model} & \multicolumn{7}{c|}{English} & \multicolumn{7}{c}{Chinese} \\
            ~ & B@1 & B@2 & B@3 & B@4 & M & R & C & B@1 & B@2 & B@3 & B@4 & M & R & C \\
        \midrule
        \multicolumn{15}{c}{\textit{Zero-shot}} \\
        \midrule
        InternVL2-1B~\cite{internvl2} & 18.47 & 8.37 & 3.87 & 1.76 & 13.43 & 19.10 & 4.89 
            & 0.01 & 0.00 & 0.00 & 0.00 & 0.03 & 0.01 & 0.00 \\
        InternVL2-2B~\cite{internvl2} & 30.13 & 13.87 & 6.74 & 3.35 & 11.96 & 22.20 & 20.52 
            & 25.28 & 8.84 & 3.64 & 1.19 & 15.77 & 25.93 & 11.84 \\
        InternVL2-8B~\cite{internvl2} & 29.96 & 14.33 & 7.05 & 3.41 & 12.80 & 22.36 & 21.17 
            & 29.03 & 10.37 & 4.64 & 1.78 & 14.01 & 26.92 & 13.96 \\
        DeepseekVL-1.3B~\cite{lu2024deepseek} & 29.08 & 14.16 & 7.08 & 3.53 & 12.92 & 22.92 & 22.37 
            & 0.12 & 0.04 & 0.02 & 0.01 & 0.53 & 2.06 & 2.33 \\
        DeepseekVL-7B~\cite{lu2024deepseek} & 28.99 & 13.49 & 6.68 & 3.31 & 13.09 & 22.58 & 20.70 
            & 24.18 & 8.69 & 3.84 & 1.46 & 11.94 & 19.84 & 14.54 \\
        LLaVA-1.5-7B~\cite{llava} & \textbf{31.70} & \textbf{16.06} & \textbf{8.48} & \textbf{4.15} & \textbf{14.30} & \textbf{24.47} & \textbf{27.76} 
            & 13.48 & 5.02 & 2.11 & 0.81 & 13.47 & 26.51 & 11.25 \\
        Qwen2-VL-2B~\cite{wang2024qwen2} & 29.26 & 14.22 & 6.99 & 3.41 & 11.81 & 20.73 & 21.39 
            & 27.99 & 10.67 & 4.67 & 2.40 & 15.07 & 23.91 & 15.84 \\
        Qwen2-VL-7B~\cite{wang2024qwen2} & 31.68 & 15.26 & 7.52 & 3.38 & 13.74 & 24.07 & 24.26 
            & \textbf{36.75} & \textbf{15.21} & \textbf{7.10} & \textbf{3.67} & \textbf{16.64} & \textbf{31.67} & \textbf{23.39} \\
        \midrule
    
        \multicolumn{15}{c}{\textit{Supervised Fine-tuning}} \\
        \midrule
        InternVL2-1B~\cite{internvl2} & 74.13 & 62.41 & 52.91 & 45.19 & 38.98 & 66.99 & 253.76 
            & 65.42 & 49.95 & 39.58 & 31.84 & 35.57 & 59.42 & 157.52 \\
        InternVL2-2B~\cite{internvl2} & 75.23 & 63.52 & 54.40 & 47.08 & 37.33 & 65.90 & 253.59 
            & 66.35 & 50.22 & 39.43 & 31.18 & 35.43 & 59.47 & 158.72 \\
        InternVL2-8B~\cite{internvl2} & 76.40 & 65.00 & 55.96 & 48.48 & 39.33 & 67.19 & 266.22 
            & 65.59 & 50.38 & 40.25 & 32.57 & 34.38 & 59.30 & 152.49 \\
        DeepseekVL-1.3B~\cite{lu2024deepseek} & 75.63 & 64.19 & 54.94 & 47.31 & 39.56 & 68.03 & 270.01 
            & 67.88 & 52.35 & 41.97 & 33.97 & 36.69 & 61.24 & 167.64 \\
        DeepseekVL-7B~\cite{lu2024deepseek} & \textbf{77.20} & \textbf{65.85} & 56.66 & 49.19 & 39.20 & 68.64 & 275.51 
            & \textbf{69.49} & \textbf{54.23} & \textbf{44.01} & \textbf{36.31} & \textbf{36.98} & \textbf{61.79} & \textbf{173.51} \\
        LLaVA-1.5-7B~\cite{llava} & 76.96 & 65.83 & 56.61 & 49.07 & \textbf{40.44} & \textbf{69.29} & 275.05 
            & 68.03 & 52.92 & 42.67 & 35.04 & 36.44 & 61.53 & 167.96 \\
        Qwen2-VL-2B~\cite{wang2024qwen2} & 76.86 & 65.73 & \textbf{56.89} & \textbf{49.90} & 39.34 & 68.45 & \textbf{282.15} 
            & 54.09 & 40.71 & 31.87 & 25.26 & 34.64 & 58.46 & 151.56 \\
        Qwen2-VL-7B~\cite{wang2024qwen2} & 74.67 & 62.82 & 53.48 & 46.02 & 38.95 & 66.41 & 267.61 
            & 65.84 & 50.06 & 39.78 & 32.14 & 35.24 & 59.19 & 159.89 \\
        \midrule

        \multicolumn{15}{c}{\textit{Multi-lingual Training}} \\
        \midrule
        InternVL2-1B~\cite{internvl2} & 73.35 & 61.43 & 52.04 & 44.57 & 38.89 & 66.08 & 257.55 
            & 66.96 & 51.66 & 41.77 & 34.31 & 35.64 & 60.86 & 169.67 \\
    
        InternVL2-2B~\cite{internvl2} & 72.64 & 60.64 & 51.17 & 43.73 & 38.42 & 65.52 & 245.90 
            & 66.05 & 50.19 & 39.71 & 31.86 & 35.40 & 60.05 & 162.07 \\
    
        InternVL2-8B~\cite{internvl2} & 71.29 & 59.07 & 49.37 & 41.57 & 37.70 & 63.33 & 237.63 
            & 65.97 & 50.22 & 39.99 & 32.45 & 35.46 & 59.01 & 150.52\\
    
        DeepseekVL-1.3B~\cite{lu2024deepseek} & 74.15 & 62.74 & 53.48 & 45.96 & 40.09 & 67.68 & 267.86 
            & 67.53 & \textbf{52.65} & \textbf{42.47} & \textbf{34.66} & \textbf{36.60} & \textbf{62.04} & \textbf{175.06} \\
    
        DeepseekVL-7B~\cite{lu2024deepseek} & 76.55 & 65.49 & 56.58 & \textbf{49.40} & 40.17 & 68.40 & \textbf{288.18} 
            & 66.64 & 51.34 & 40.90 & 33.08 & 36.39 & 60.43 & 159.03\\
        LLaVA-1.5-7B~\cite{llava} & \textbf{76.98} & \textbf{66.11} & \textbf{56.94} & 49.23 & \textbf{40.47} & \textbf{69.48} & 276.01 
            & 66.88 & 51.10 & 40.63 & 32.75 & 36.18 & 60.16 & 163.06\\
    
        Qwen2-VL-2B~\cite{wang2024qwen2} & 76.40 & 64.54 & 54.96 & 47.15 & 39.61 & 68.39 & 274.29 
            & 65.16 & 49.61 & 39.10 & 31.11 & 35.76 & 59.74 & 161.11 \\
    
        Qwen2-VL-7B~\cite{wang2024qwen2} & 76.18 & 64.60 & 55.30 & 47.78 & 38.88 & 67.07 & 267.05 
            & \textbf{67.72} & 51.57 & 40.84 & 32.69 & 36.22 & 60.63 & 161.18\\
        \bottomrule
    \end{tabular}
    \end{table*}

\begin{table*}
    \caption{Comparison of cross-dataset transfer results of Qwen2-VL-7B and MG on BRSIC.}
    \label{tab:cross-data}
    \centering
    \setlength{\tabcolsep}{3.5pt}
    \begin{tabular}{@{}ll|ccccccc|ccccccc@{}}
    \toprule
    \multicolumn{2}{l|}{\multirow{2}{*}{Transfer Setting}} & \multicolumn{7}{c|}{English} & \multicolumn{7}{c}{Chinese} \\
    
    & & B@1 & B@2 & B@3 & B@4 & M & R & C & B@1 & B@2 & B@3 & B@4 & M & R & C \\
    \midrule

    \multicolumn{16}{@{}c}{\textit{Target Dataset: Sydney}} \\
    \midrule
    Zero-shot & Qwen2-VL-7B & 37.70 & 16.36 & 8.40 & 4.77 & 14.68 & 32.94 & 19.37 & 40.48 & 20.36 & 11.83 & 5.57 & 17.92 & 34.91 & 15.95 \\
    \multirow{2}{*}{UCM$\rightarrow$Sydney} 
    & Qwen2-VL-7B & 41.74 & 27.34 & 17.30 & 6.63 & 16.36 & \textbf{36.97} & 20.21 
        & \textbf{52.89} & \textbf{33.01} & 20.45 & 11.56 & \textbf{22.11} & \textbf{43.86} & \textbf{28.88} \\
        
    & MG & 42.85 & \textbf{30.78} & \textbf{18.23} & \textbf{12.01} & \textbf{16.41} & 36.41 & 25.17 
        & 39.84 & 28.98 & \textbf{21.78} & \textbf{16.54} & 16.76 & 33.23 & 27.89 \\

    \multirow{2}{*}{RSICD$\rightarrow$Sydney}
    & Qwen2-VL-7B & 48.07 & 23.29 & 13.27 & 7.37 & 16.10 & 27.84 & 33.10 
        & 42.46 & 23.47 & 16.47 & 12.19 & 17.04 & 33.84 & 20.89 \\
    & MG & \textbf{49.78} & 24.57 & 14.10 & 7.99 & 16.36 & 30.91 & \textbf{33.30}
        & 36.43 & 19.35 & 11.90 & 7.63 & 15.41 & 27.54 & 19.02 \\
    
    \midrule
    
    \multicolumn{16}{@{}c}{\textit{Target Dataset: UCM}} \\
    \midrule
    Zero-shot & Qwen2-VL-7B & 36.65 & 19.76 & 12.07 & 7.53 & 15.90 & 32.71 & 45.66 
        & 45.85 & 26.83 & 16.60 & 10.33 & 21.16 & 40.30 & 44.97 \\
    \multirow{2}{*}{Sydney$\rightarrow$UCM}
    & Qwen2-VL-7B & \textbf{50.66} & \textbf{34.38} & \textbf{24.66} & \textbf{17.95} & \textbf{22.35} & \textbf{42.40} & \textbf{73.93} 
        & \textbf{53.42} & \textbf{34.36} & \textbf{23.94} & \textbf{17.92} & \textbf{26.57} & \textbf{46.32} & \textbf{59.87} \\

    & MG & 44.19 & 26.31 & 16.21 & 7.52 & 14.15 & 34.75 & 33.69 
        & 44.55 & 27.07 & 18.08 & 12.57 & 18.57 & 33.28 & 44.40 \\

    \multirow{2}{*}{RSICD$\rightarrow$UCM}
    & Qwen2-VL-7B & 46.52 & 25.84 & 16.09 & 10.20 & 17.97 & 34.51 & 62.30 
        & 46.83 & 26.59 & 16.53 & 10.53 & 20.78 & 39.56 & 54.46 \\

    & MG & 44.53 & 24.63 & 15.17 & 9.82 & 15.85 & 32.38 & 59.14 
        & 35.03 & 24.12 & 17.52 & 12.62 & 16.30 & 31.80 & 53.49 \\
    
    \midrule
    
    \multicolumn{16}{@{}c}{\textit{Target Dataset: RSICD}} \\
    \midrule
    Zero-shot & Qwen2-VL-7B & 31.68 & 15.26 & 7.52 & 3.38 & 13.74 & 24.07 & 24.26 & 36.75 & 15.21 & 7.10 & 3.67 & 16.64 & 31.67 & 23.39 \\
    \multirow{2}{*}{UCM$\rightarrow$RSICD}
    & Qwen2-VL-7B & 40.46 & \textbf{22.64} & \textbf{13.31} & \textbf{7.78} & 15.55 & \textbf{30.58} & 42.61 & \textbf{38.34} & \textbf{18.44} & \textbf{9.90} & \textbf{5.42} & \textbf{19.65} & \textbf{34.93} & \textbf{31.04} \\
    & MG & 35.08 & 18.79 & 10.83 & 6.17 & 12.01 & 25.55 & 31.09 & 29.20 & 15.46 & 8.34 & 4.08 & 13.60 & 22.27 & 24.33 \\
    \multirow{2}{*}{Sydney$\rightarrow$RSICD}
    & Qwen2-VL-7B & \textbf{40.50} & 22.06 & 12.13 & 6.18 & \textbf{16.61} & 27.87 & \textbf{45.93} & 37.09 & 17.53 & 8.94 & 4.84 & 18.89 & 32.94 & 22.35 \\
    & MG & 31.13 & 11.56 & 4.17 & 1.17 & 10.79 & 22.21 & 9.90 & 26.74 & 10.96 & 4.74 & 2.25 & 12.99 & 19.56 & 12.39 \\
    \bottomrule
    \end{tabular}
    \end{table*}

We conduct a comprehensive evaluation of eight state-of-the-art LVLMs, including InternVL2~\cite{internvl2}, Qwen2-VL~\cite{wang2024qwen2}, LLaVA~\cite{llava}, and DeepseekVL~\cite{lu2024deepseek} with 1.3B, 2B, 7B, and 8B parameters, across multiple experimental settings. Our analysis spans zero-shot inference, supervised fine-tuning, and multi-lingual training scenarios, revealing the models' cross-lingual capabilities and adaptation patterns. The results demonstrate that while these models show promising potential in cross-lingual remote sensing image captioning, their performance varies significantly based on model scale, training strategy, and target language, with notable improvements achieved through supervised fine-tuning and multi-lingual training approaches.

\subsubsection{Zero-shot Performance}
The zero-shot evaluation reveals significant disparities in cross-lingual generalization capabilities. As shown in Tables~\ref{tab:sydney},~\ref{tab:ucm}, and~\ref{tab:rsicd}, models demonstrate varying degrees of success in English caption generation, with BLEU@1 scores ranging from 18.47 to 39.97. LLaVA-1.5-7B consistently achieves the best English zero-shot performance across datasets (BLEU@1: 39.97 on Sydney, 37.61 on RSICD). For Chinese caption generation, most smaller models (1-2B parameters) struggle significantly, often producing no meaningful captions. However, larger models like Qwen2-VL-7B show promising Chinese zero-shot capabilities, achieving the highest scores across all datasets (BLEU@1: 40.48 on Sydney, 45.85 on UCM, 36.75 on RSICD).

The performance disparity between English and Chinese can be attributed to several factors: the dominance of English in pre-training data, the complexity of Chinese sentence structures, and the limited incorporation of remote sensing domain knowledge in Chinese during pre-training.

\subsubsection{Supervised Fine-tuning Results}
Supervised fine-tuning significantly improves performance across both languages. For English captions, most models achieve BLEU@1 scores above 74, with DeepseekVL-7B and LLaVA-1.5-7B consistently ranking among the top performers. DeepseekVL-7B achieves the highest scores on RSICD (BLEU@1: 77.20), while LLaVA-1.5-7B excels on UCM (BLEU@1: 88.22).
For Chinese caption generation, the improvements through supervised fine-tuning are substantial across all models. Qwen2-VL-7B and DeepseekVL-7B demonstrate particularly strong cross-lingual capabilities, with Qwen2-VL-7B achieving exceptional results on Sydney (BLEU@1: 77.12, CIDEr: 251.52) and UCM (BLEU@1: 80.07, CIDEr: 257.42), while DeepseekVL-7B leads on RSICD (BLEU@1: 69.49, CIDEr: 173.51). 

\subsubsection{Multi-lingual Training Analysis}

Multi-lingual training refers to the joint training process where models are trained on mixed English and Chinese samples, responding to language-specific prompts to generate captions in the corresponding language.
It reveals interesting patterns in model behavior. For English captions, performance remains robust, with some models even showing improvements over single-language fine-tuning. InternVL2-8B achieves remarkable results on Sydney (BLEU@1: 79.89), while Qwen2-VL-7B excels on UCM (BLEU@1: 87.36, CIDEr: 376.71).
The Chinese caption generation benefits substantially from multi-lingual training, with more balanced performance across languages. DeepseekVL-1.3B shows surprisingly strong results despite its smaller size, particularly on RSICD (BLEU@1: 67.53, CIDEr: 175.06). This suggests that model size is not the sole determinant of multi-lingual capability.

\subsubsection{Discussion and Analysis}
Our comprehensive evaluation yields several significant insights into multi-lingual vision-language learning. First, we observe a strong correlation between model scale and multi-lingual performance, with larger models (7B-8B parameters) exhibiting superior zero-shot capabilities. Nevertheless, it is noteworthy that smaller architectures, exemplified by DeepseekVL-1.3B, can achieve comparable results through well-designed multi-lingual training strategies.
Furthermore, our analysis reveals an inherent asymmetry in language performance. Although English generation consistently outperforms Chinese, this disparity diminishes substantially through supervised fine-tuning and multi-lingual training, indicating that pre-training biases can be effectively addressed through targeted optimization techniques.
Architectural design emerges as a crucial factor in multi-lingual generalization. Our experiments demonstrate that different architectures exhibit distinct strengths - DeepseekVL achieves remarkable balance in cross-lingual performance, while Qwen2-VL demonstrates exceptional zero-shot capabilities. These findings emphasize the significant role of architectural decisions in determining a model's cross-lingual generalization capacity.

These findings point to several future research directions: developing more efficient multi/cross-lingual adaptation techniques, investigating domain-specific pre-training strategies for remote sensing applications, and exploring balanced multi-lingual training approaches that can maintain consistent performance across languages while reducing computational requirements.

\subsection{Cross-dataset Transfer Analysis}
Cross-dataset transfer refers to the process of transferring a trained model from one dataset to another.
We compare Qwen2-VL-7B, a large vision-language model, with MG, a traditional vision-language model, to understand how model scale affects transfer capabilities.

The transfer approach demonstrates consistent improvements over zero-shot baselines across all target datasets. When targeting Sydney dataset, UCM$\rightarrow$Sydney transfer using Qwen2-VL-7B improves BLEU@1 scores from 37.70 to 41.74 in English. Notably, MG achieves comparable or even slightly better performance in this setting (BLEU@1: 42.85), suggesting that traditional models remain competitive for transfers between similar datasets.

The effectiveness of transfer learning varies based on both dataset characteristics and model architecture. For transfers between visually similar datasets (UCM$\rightarrow$Sydney), both models show distinct strengths: Qwen2-VL-7B achieves higher overall performance in Chinese caption generation (BLEU@1: 52.89) compared to MG (39.84), while MG demonstrates better precision in English generation (B@4: 12.01 vs 6.63). In scenarios involving larger-to-smaller dataset transfers (RSICD$\rightarrow$Sydney/UCM), Qwen2-VL-7B exhibits more robust performance in both languages, particularly in maintaining semantic consistency as evidenced by higher METEOR and ROUGE scores.

Notably, Qwen2-VL-7B demonstrates remarkable zero-shot capabilities in Chinese caption generation, even surpassing the cross-dataset transfer performance of traditional models on certain metrics. For instance, its zero-shot Chinese performance on UCM (BLEU@1: 45.85) outperforms MG's transfer learning results from Sydney (BLEU@1: 44.55) and RSICD (BLEU@1: 35.03). This suggests that large vision-language models have acquired strong multilingual understanding during pre-training that can sometimes exceed the benefits of transfer learning in traditional models.

\section{Conclusion}
This paper presents BRSIC, the first bilingual benchmark for remote sensing image captioning, along with comprehensive evaluations of both traditional vision-language models and large vision-language models across languages.
The construction and analysis of BRSIC, comprising 13,634 images and 68,170 parallel English-Chinese caption pairs across three widely-used RSIC datasets, reveals important insights into multi-lingual RSIC challenges.
Our systematic evaluation framework addresses critical dataset distribution mismatches and provides a reliable foundation for assessing model capabilities across languages.
The comprehensive assessment of state-of-the-art large vision-language models through zero-shot inference, supervised fine-tuning, and cross-dataset transfer reveals both the inherent advantages of larger models and the effectiveness of targeted training strategies in improving multi-lingual performance.

\section{Acknowledgements}
The authors acknowledge the use of Claude-3.5-Sonnet for grammar refinement and language editing. All AI-generated content is reviewed and corrected by the authors.

\bibliographystyle{IEEEtran}
\bibliography{ref}

\begin{thebibliography}{10}
\providecommand{\url}[1]{#1}
\csname url@samestyle\endcsname
\providecommand{\newblock}{\relax}
\providecommand{\bibinfo}[2]{#2}
\providecommand{\BIBentrySTDinterwordspacing}{\spaceskip=0pt\relax}
\providecommand{\BIBentryALTinterwordstretchfactor}{4}
\providecommand{\BIBentryALTinterwordspacing}{\spaceskip=\fontdimen2\font plus
\BIBentryALTinterwordstretchfactor\fontdimen3\font minus \fontdimen4\font\relax}
\providecommand{\BIBforeignlanguage}[2]{{%
\expandafter\ifx\csname l@#1\endcsname\relax
\typeout{** WARNING: IEEEtran.bst: No hyphenation pattern has been}%
\typeout{** loaded for the language `#1'. Using the pattern for}%
\typeout{** the default language instead.}%
\else
\language=\csname l@#1\endcsname
\fi
#2}}
\providecommand{\BIBdecl}{\relax}
\BIBdecl

\bibitem{4215085}
J.-Y. Rau, L.-C. Chen, J.-K. Liu, and T.-H. Wu, ``Dynamics monitoring and disaster assessment for watershed management using time-series satellite images,'' \emph{IEEE Transactions on Geoscience and Remote Sensing}, vol.~45, no.~6, pp. 1641--1649, 2007.

\bibitem{sen}
Q.~Zhou, J.~Gao, Y.~Yuan, and Q.~Wang, ``Single-stream extractor network with contrastive pre-training for remote-sensing change captioning,'' \emph{IEEE Transactions on Geoscience and Remote Sensing}, vol.~62, pp. 1--14, 2024.

\bibitem{885202}
M.~Imhoff, C.~Tucker, W.~Lawrence, and D.~Stutzer, ``The use of multisource satellite and geospatial data to study the effect of urbanization on primary productivity in the united states,'' \emph{IEEE Transactions on Geoscience and Remote Sensing}, vol.~38, no.~6, pp. 2549--2556, 2000.

\bibitem{9308980}
Q.~Wang, W.~Huang, X.~Zhang, and X.~Li, ``Word–sentence framework for remote sensing image captioning,'' \emph{IEEE Transactions on Geoscience and Remote Sensing}, vol.~59, no.~12, pp. 10\,532--10\,543, 2021.

\bibitem{9153154}
X.~Li, X.~Zhang, W.~Huang, and Q.~Wang, ``Truncation cross entropy loss for remote sensing image captioning,'' \emph{IEEE Transactions on Geoscience and Remote Sensing}, vol.~59, no.~6, pp. 5246--5257, 2021.

\bibitem{jiang2018learning}
W.~Jiang, L.~Ma, X.~Chen, H.~Zhang, and W.~Liu, ``Learning to guide decoding for image captioning,'' in \emph{Proceedings of the AAAI Conference on Artificial Intelligence}, vol.~32, no.~1, 2018.

\bibitem{yao2017boosting}
T.~Yao, Y.~Pan, Y.~Li, Z.~Qiu, and T.~Mei, ``Boosting image captioning with attributes,'' in \emph{Proceedings of the IEEE international conference on computer vision}, 2017, pp. 4894--4902.

\bibitem{qu2016deep}
B.~Qu, X.~Li, D.~Tao, and X.~Lu, ``Deep semantic understanding of high resolution remote sensing image,'' in \emph{2016 International conference on computer, information and telecommunication systems (Cits)}.\hskip 1em plus 0.5em minus 0.4em\relax IEEE, 2016, pp. 1--5.

\bibitem{mlat}
C.~Liu, R.~Zhao, and Z.~Shi, ``Remote-sensing image captioning based on multilayer aggregated transformer,'' \emph{IEEE Geoscience and Remote Sensing Letters}, vol.~19, pp. 1--5, 2022.

\bibitem{MG}
L.~Meng, J.~Wang, R.~Meng, Y.~Yang, and L.~Xiao, ``A multiscale grouping transformer with clip latents for remote sensing image captioning,'' \emph{IEEE Transactions on Geoscience and Remote Sensing}, vol.~62, pp. 1--15, 2024.

\bibitem{bui2023uit}
D.~C. Bui, N.~H. Nguyen, and K.~Nguyen, ``Uit-openviic: A novel benchmark for evaluating image captioning in vietnamese,'' \emph{arXiv preprint arXiv:2305.04166}, 2023.

\bibitem{aoanet}
L.~Huang, W.~Wang, J.~Chen, and X.-Y. Wei, ``Attention on attention for image captioning,'' in \emph{Proceedings of the IEEE/CVF international conference on computer vision}, 2019, pp. 4634--4643.

\bibitem{lu2019sound}
X.~Lu, B.~Wang, and X.~Zheng, ``Sound active attention framework for remote sensing image captioning,'' \emph{IEEE Transactions on Geoscience and Remote Sensing}, vol.~58, no.~3, pp. 1985--2000, 2019.

\bibitem{zhang2019description}
X.~Zhang, X.~Wang, X.~Tang, H.~Zhou, and C.~Li, ``Description generation for remote sensing images using attribute attention mechanism,'' \emph{Remote Sensing}, vol.~11, no.~6, p. 612, 2019.

\bibitem{zhang2019lam}
Z.~Zhang, W.~Diao, W.~Zhang, M.~Yan, X.~Gao, and X.~Sun, ``Lam: Remote sensing image captioning with label-attention mechanism,'' \emph{Remote Sensing}, vol.~11, no.~20, p. 2349, 2019.

\bibitem{liu2022remote}
C.~Liu, R.~Zhao, and Z.~Shi, ``Remote-sensing image captioning based on multilayer aggregated transformer,'' \emph{IEEE Geoscience and Remote Sensing Letters}, vol.~19, pp. 1--5, 2022.

\bibitem{Rscama}
C.~Liu, K.~Chen, B.~Chen, H.~Zhang, Z.~Zou, and Z.~Shi, ``Rscama: Remote sensing image change captioning with state space model,'' \emph{IEEE Geoscience and Remote Sensing Letters}, vol.~21, pp. 1--5, 2024.

\bibitem{gu2023mamba}
A.~Gu and T.~Dao, ``Mamba: Linear-time sequence modeling with selective state spaces,'' \emph{arXiv preprint arXiv:2312.00752}, 2023.

\bibitem{bai2023qwen}
J.~Bai, S.~Bai, Y.~Chu, Z.~Cui, K.~Dang, X.~Deng, Y.~Fan, W.~Ge, Y.~Han, F.~Huang \emph{et~al.}, ``Qwen technical report,'' \emph{arXiv preprint arXiv:2309.16609}, 2023.

\bibitem{cai2024internlm2}
Z.~Cai, M.~Cao, H.~Chen, K.~Chen, K.~Chen, X.~Chen, X.~Chen, Z.~Chen, Z.~Chen, P.~Chu \emph{et~al.}, ``Internlm2 technical report,'' \emph{arXiv preprint arXiv:2403.17297}, 2024.

\bibitem{llama2}
H.~Touvron, L.~Martin, K.~Stone, P.~Albert, A.~Almahairi, Y.~Babaei, N.~Bashlykov, S.~Batra, P.~Bhargava, S.~Bhosale \emph{et~al.}, ``Llama 2: Open foundation and fine-tuned chat models,'' \emph{arXiv preprint arXiv:2307.09288}, 2023.

\bibitem{wang2024qwen2}
P.~Wang, S.~Bai, S.~Tan, S.~Wang, Z.~Fan, J.~Bai, K.~Chen, X.~Liu, J.~Wang, W.~Ge \emph{et~al.}, ``Qwen2-vl: Enhancing vision-language model's perception of the world at any resolution,'' \emph{arXiv preprint arXiv:2409.12191}, 2024.

\bibitem{internvl2}
Z.~Chen, W.~Wang, H.~Tian, S.~Ye, Z.~Gao, E.~Cui, W.~Tong, K.~Hu, J.~Luo, Z.~Ma \emph{et~al.}, ``How far are we to gpt-4v? closing the gap to commercial multimodal models with open-source suites,'' \emph{arXiv preprint arXiv:2404.16821}, 2024.

\bibitem{internvl2.5}
Z.~Chen, W.~Wang, Y.~Cao, Y.~Liu, Z.~Gao, E.~Cui, J.~Zhu, S.~Ye, H.~Tian, Z.~Liu \emph{et~al.}, ``Expanding performance boundaries of open-source multimodal models with model, data, and test-time scaling,'' \emph{arXiv preprint arXiv:2412.05271}, 2024.

\bibitem{lu2024deepseek}
H.~Lu, W.~Liu, B.~Zhang, B.~Wang, K.~Dong, B.~Liu, J.~Sun, T.~Ren, Z.~Li, H.~Yang \emph{et~al.}, ``Deepseek-vl: towards real-world vision-language understanding,'' \emph{arXiv preprint arXiv:2403.05525}, 2024.

\bibitem{llava}
H.~Liu, C.~Li, Y.~Li, and Y.~J. Lee, ``Improved baselines with visual instruction tuning,'' in \emph{Proceedings of the IEEE/CVF Conference on Computer Vision and Pattern Recognition}, 2024, pp. 26\,296--26\,306.

\bibitem{lu2017exploring}
X.~Lu, B.~Wang, X.~Zheng, and X.~Li, ``Exploring models and data for remote sensing image caption generation,'' \emph{IEEE Transactions on Geoscience and Remote Sensing}, vol.~56, no.~4, pp. 2183--2195, 2017.

\bibitem{nwpu-caption}
Q.~Cheng, H.~Huang, Y.~Xu, Y.~Zhou, H.~Li, and Z.~Wang, ``Nwpu-captions dataset and mlca-net for remote sensing image captioning,'' \emph{IEEE Transactions on Geoscience and Remote Sensing}, vol.~60, pp. 1--19, 2022.

\bibitem{li2020multi}
Y.~Li, S.~Fang, L.~Jiao, R.~Liu, and R.~Shang, ``A multi-level attention model for remote sensing image captions,'' \emph{Remote Sensing}, vol.~12, no.~6, p. 939, 2020.

\bibitem{lan2017fluency}
W.~Lan, X.~Li, and J.~Dong, ``Fluency-guided cross-lingual image captioning,'' in \emph{Proceedings of the 25th ACM international conference on Multimedia}, 2017, pp. 1549--1557.

\bibitem{li2019coco}
X.~Li, C.~Xu, X.~Wang, W.~Lan, Z.~Jia, G.~Yang, and J.~Xu, ``Coco-cn for cross-lingual image tagging, captioning, and retrieval,'' \emph{IEEE Transactions on Multimedia}, vol.~21, no.~9, pp. 2347--2360, 2019.

\bibitem{hitschler2016multimodal}
J.~Hitschler, S.~Schamoni, and S.~Riezler, ``Multimodal pivots for image caption translation,'' \emph{arXiv preprint arXiv:1601.03916}, 2016.

\bibitem{pham2024ktvic}
A.-C. Pham, V.-Q. Nguyen, T.-H. Vuong, and Q.-T. Ha, ``Ktvic: A vietnamese image captioning dataset on the life domain,'' \emph{arXiv preprint arXiv:2401.08100}, 2024.

\bibitem{DBLP:conf/cvpr/ZhouZW0LYL21}
M.~Zhou, L.~Zhou, S.~Wang, Y.~Cheng, L.~Li, Z.~Yu, and J.~Liu, ``{UC2:} universal cross-lingual cross-modal vision-and-language pre-training,'' in \emph{{IEEE} Conference on Computer Vision and Pattern Recognition, {CVPR} 2021, virtual, June 19-25, 2021}.\hskip 1em plus 0.5em minus 0.4em\relax Computer Vision Foundation / {IEEE}, 2021, pp. 4155--4165.

\bibitem{Tsutsui2017UsingAT}
S.~Tsutsui and D.~J. Crandall, ``Using artificial tokens to control languages for multilingual image caption generation,'' \emph{ArXiv}, vol. abs/1706.06275, 2017.

\bibitem{Xiao2020AnIM}
Y.~Xiao and T.~Lu, ``An improved method of cross-lingual image captioning based on fluency-guided,'' \emph{2020 5th International Conference on Control, Robotics and Cybernetics (CRC)}, pp. 165--170, 2020.

\bibitem{vinyals2015show}
O.~Vinyals, A.~Toshev, S.~Bengio, and D.~Erhan, ``Show and tell: A neural image caption generator,'' in \emph{Proceedings of the IEEE conference on computer vision and pattern recognition}, 2015, pp. 3156--3164.

\bibitem{yoshikawa2017stair}
Y.~Yoshikawa, Y.~Shigeto, and A.~Takeuchi, ``Stair captions: Constructing a large-scale japanese image caption dataset,'' \emph{arXiv preprint arXiv:1705.00823}, 2017.

\bibitem{elliott2016multi30k}
D.~Elliott, S.~Frank, K.~Sima'an, and L.~Specia, ``Multi30k: Multilingual english-german image descriptions,'' in \emph{5th Workshop on Vision and Language}.\hskip 1em plus 0.5em minus 0.4em\relax Association for Computational Linguistics (ACL), 2016, pp. 70--74.

\bibitem{miyazaki2016cross}
T.~Miyazaki and N.~Shimizu, ``Cross-lingual image caption generation,'' in \emph{Proceedings of the 54th Annual Meeting of the Association for Computational Linguistics (Volume 1: Long Papers)}, 2016, pp. 1780--1790.

\bibitem{zhao2021high}
R.~Zhao, Z.~Shi, and Z.~Zou, ``High-resolution remote sensing image captioning based on structured attention,'' \emph{IEEE Transactions on Geoscience and Remote Sensing}, vol.~60, pp. 1--14, 2021.

\bibitem{wang2020word}
Q.~Wang, W.~Huang, X.~Zhang, and X.~Li, ``Word--sentence framework for remote sensing image captioning,'' \emph{IEEE Transactions on Geoscience and Remote Sensing}, vol.~59, no.~12, pp. 10\,532--10\,543, 2020.

\bibitem{shen2020remote}
X.~Shen, B.~Liu, Y.~Zhou, J.~Zhao, and M.~Liu, ``Remote sensing image captioning via variational autoencoder and reinforcement learning,'' \emph{Knowledge-Based Systems}, vol. 203, p. 105920, 2020.

\bibitem{li2024learning}
Y.~Li, X.~Zhang, X.~Cheng, X.~Tang, and L.~Jiao, ``Learning consensus-aware semantic knowledge for remote sensing image captioning,'' \emph{Pattern Recognition}, vol. 145, p. 109893, 2024.

\bibitem{yao2024minicpm}
Y.~Yao, T.~Yu, A.~Zhang, C.~Wang, J.~Cui, H.~Zhu, T.~Cai, H.~Li, W.~Zhao, Z.~He \emph{et~al.}, ``Minicpm-v: A gpt-4v level mllm on your phone,'' \emph{arXiv preprint arXiv:2408.01800}, 2024.

\bibitem{radford2021learning}
A.~Radford, J.~W. Kim, C.~Hallacy, A.~Ramesh, G.~Goh, S.~Agarwal, G.~Sastry, A.~Askell, P.~Mishkin, J.~Clark \emph{et~al.}, ``Learning transferable visual models from natural language supervision,'' in \emph{International conference on machine learning}.\hskip 1em plus 0.5em minus 0.4em\relax PMLR, 2021, pp. 8748--8763.

\bibitem{yang2023dawn}
Z.~Yang, L.~Li, K.~Lin, J.~Wang, C.-C. Lin, Z.~Liu, and L.~Wang, ``The dawn of lmms: Preliminary explorations with gpt-4v (ision),'' \emph{arXiv preprint arXiv:2309.17421}, vol.~9, no.~1, p.~1, 2023.

\bibitem{hu2023rsgpt}
Y.~Hu, J.~Yuan, C.~Wen, X.~Lu, and X.~Li, ``Rsgpt: A remote sensing vision language model and benchmark,'' \emph{arXiv preprint arXiv:2307.15266}, 2023.

\bibitem{kuckreja2024geochat}
K.~Kuckreja, M.~S. Danish, M.~Naseer, A.~Das, S.~Khan, and F.~S. Khan, ``Geochat: Grounded large vision-language model for remote sensing,'' in \emph{Proceedings of the IEEE/CVF Conference on Computer Vision and Pattern Recognition}, 2024, pp. 27\,831--27\,840.

\bibitem{zhan2024skyeyegpt}
Y.~Zhan, Z.~Xiong, and Y.~Yuan, ``Skyeyegpt: Unifying remote sensing vision-language tasks via instruction tuning with large language model,'' \emph{arXiv preprint arXiv:2401.09712}, 2024.

\bibitem{butd}
P.~Anderson, X.~He, C.~Buehler, D.~Teney, M.~Johnson, S.~Gould, and L.~Zhang, ``Bottom-up and top-down attention for image captioning and visual question answering,'' in \emph{Proceedings of the IEEE conference on computer vision and pattern recognition}, 2018, pp. 6077--6086.

\bibitem{BLEU}
K.~Papineni, S.~Roukos, T.~Ward, and W.-J. Zhu, ``Bleu: a method for automatic evaluation of machine translation,'' \emph{Proceedings of the Association for Computational Linguistics}, pp. 311--318, 2002.

\bibitem{METEOR}
M.~Denkowski and A.~Lavie, ``Meteor universal: Language specific translation evaluation for any target language,'' \emph{Proceedings of the Workshop on Statistical Machine Translation}, pp. 376--380, 2014.

\bibitem{ROUGE_L}
G.~Doddington, ``Automatic evaluation of machine translation quality using n-gram co-occurrence statistics,'' \emph{Proceedings of the International Conference on Human Language Technology Research}, pp. 138--145, 2002.

\bibitem{CIDEr}
R.~Vedantam, C.~Lawrence~Zitnick, and D.~Parikh, ``Cider: Consensus-based image description evaluation,'' \emph{Proceedings of the IEEE Conference on Computer Vision and Pattern Recognition}, pp. 4566--4575, 2015.

\bibitem{hu2022lora}
E.~J. Hu, yelong shen, P.~Wallis, Z.~Allen-Zhu, Y.~Li, S.~Wang, L.~Wang, and W.~Chen, ``Lo{RA}: Low-rank adaptation of large language models,'' in \emph{International Conference on Learning Representations}, 2022.

\end{thebibliography}

\begin{IEEEbiography}[{\includegraphics[width=1in,height=1.25in,clip,keepaspectratio]{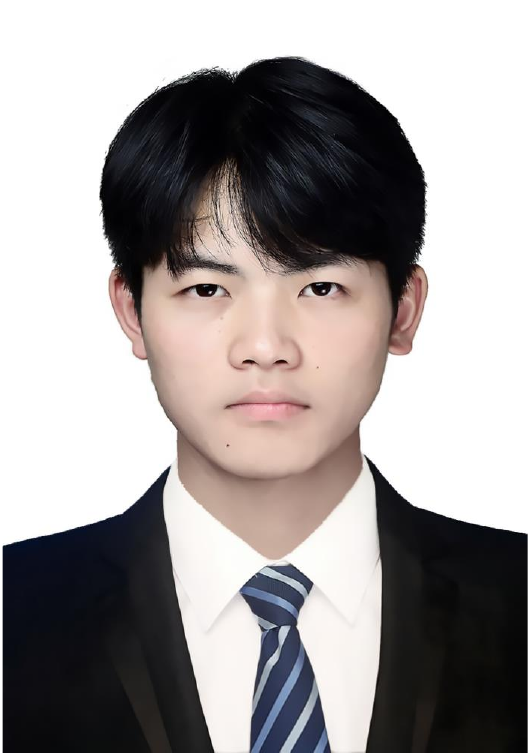}}]{Qing Zhou} is currently pursuing the Ph.D degree in computer science and technology with the school of Artificial Intelligence, Optics and Electronics (iOPEN). His research interests include computer vision and pattern recognition.
\end{IEEEbiography}

\begin{IEEEbiography}[{\includegraphics[width=1in,height=1.25in,clip,keepaspectratio]{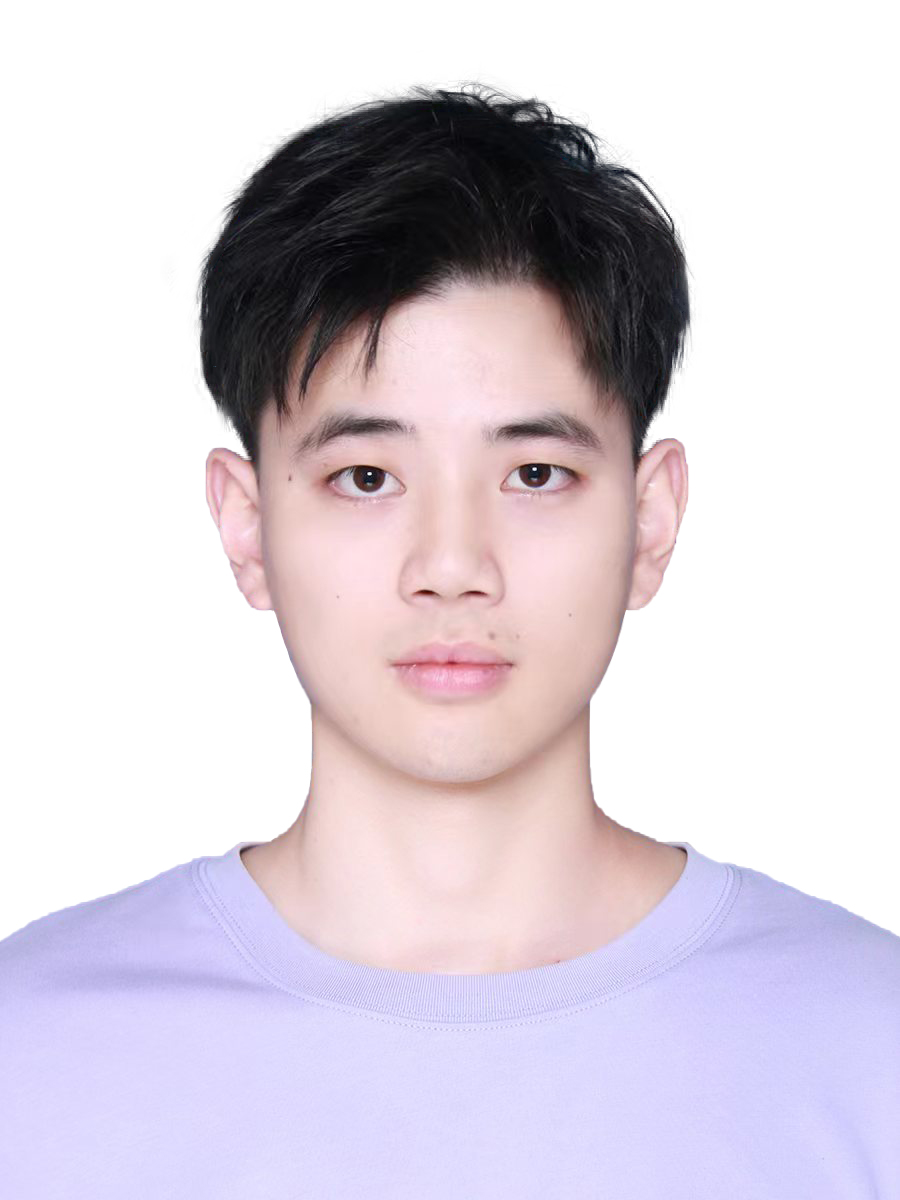}}]{Tao Yang} is currently pursuing the Ph.D degree in computer science and technology with the school of Artificial Intelligence, Optics and Electronics (iOPEN). His research interests include computer vision and pattern recognition.
\end{IEEEbiography}

\begin{IEEEbiography}[{\includegraphics[width=1in,height=1.25in,clip,keepaspectratio]{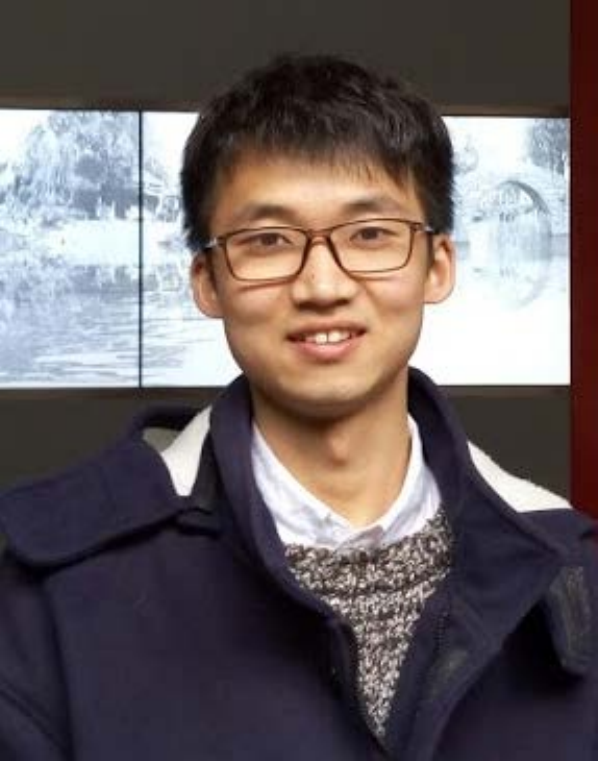}}]{Junyu Gao} (Member, IEEE) received the B.E. and Ph.D. degrees in computer science and technology from Northwestern Polytechnical University, Xi'an, China, in 2015 and 2021, respectively. He is curently a Researcher with the School of Artificial Intelligence, Optics and Electronics (iOPEN), Northwestern Polytechnical University. His research interests include computer vision and pattern recognition.
\end{IEEEbiography}

\begin{IEEEbiography}[{\includegraphics[width=1in,height=1.25in,clip,keepaspectratio]{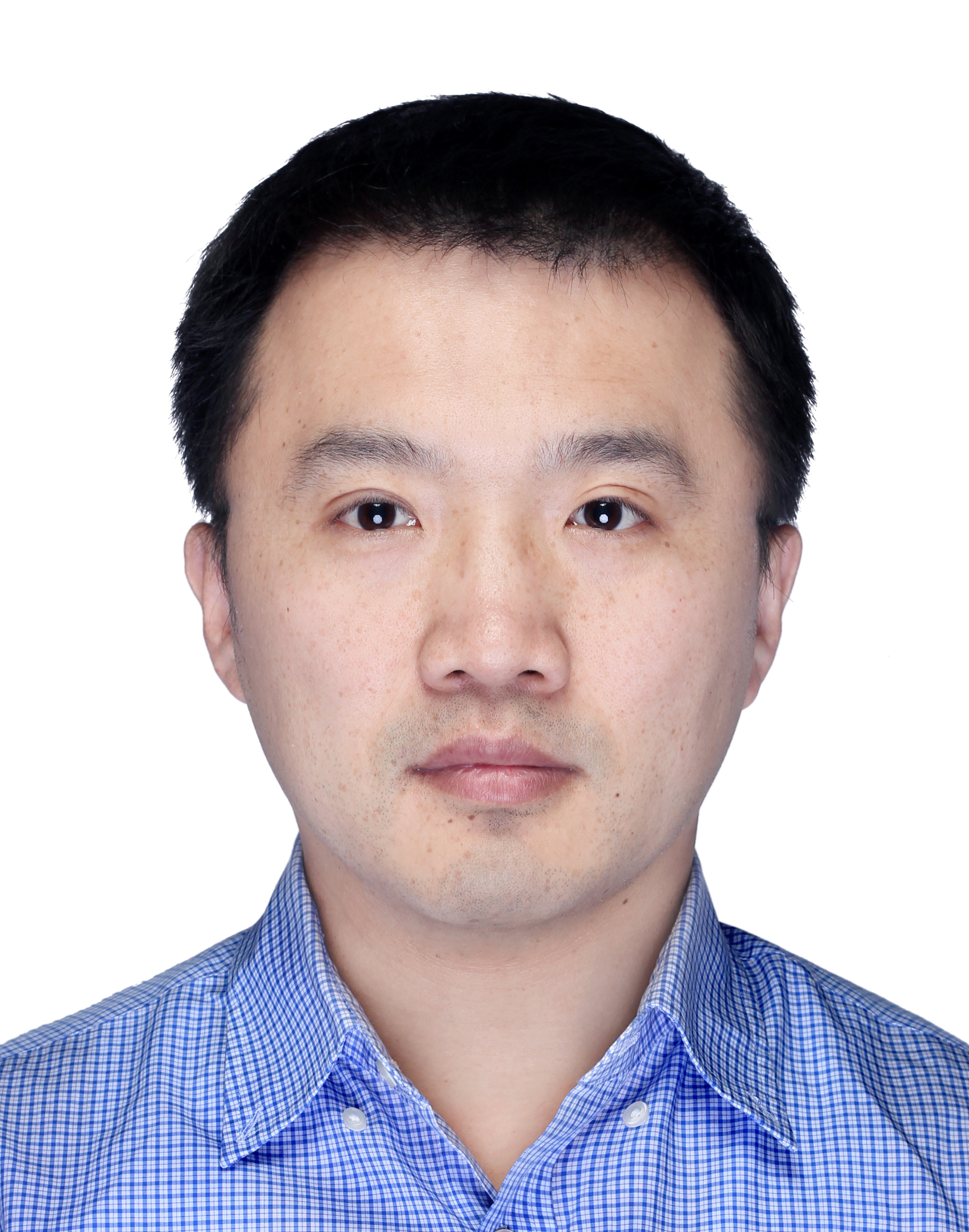}}]{Weiping Ni} received the B.S. degree from the University of Science and Technology of China, Hefei, China, in 2004, the M.S. degree from the National University of Defense Technology, Changsha, China, in 2006, and the Ph.D. degree in pattern recognition and intelligent system from Xidian University, Xi'an, China, in 2016. Since 2014, he has been a Research Associate with the Northwest Institute of Nuclear Technology, Xi'an. His research interests include remote sensing image processing and automatic target recognition.
\end{IEEEbiography}

\begin{IEEEbiography}[{\includegraphics[width=1in,height=1.25in,clip,keepaspectratio]{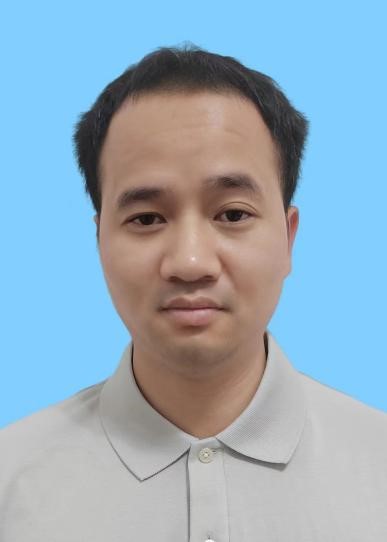}}]{Junzheng Wu} received the B.Sc. degree in automation from Tsinghua University, Beijing, China, in 2008, the M. Sc. degree in signal and information processing from Northwest Institute of Nuclear Technology, Xi'an, China, in 2011, and the Ph.D. in information and communication engineering from National University of Defense Technology, Changsha, China, in 2022, respectively. Currently, he is an Associate researcher with the department of remote sensing, Northwest Institute of Nuclear Technology, Xi'an, China. His research interests include processing of remote sensing images and machine learning.
\end{IEEEbiography}

\begin{IEEEbiography}[{\includegraphics[width=1in,height=1.25in,clip,keepaspectratio]{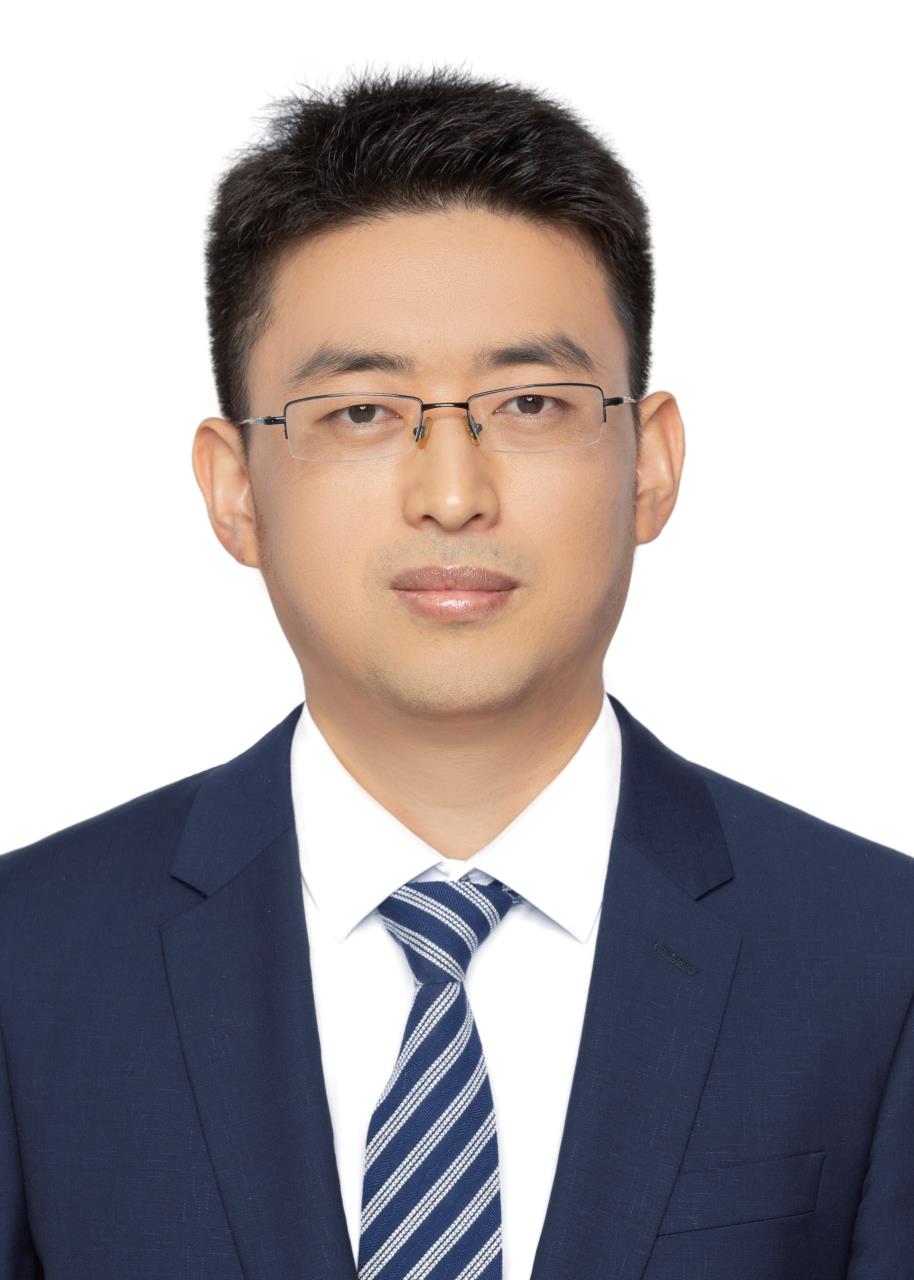}}]{Qi Wang}  (Senior Member, IEEE) received the B.E. degree in automation and the Ph.D. degree in pattern recognition and intelligent systems from the University of Science and Technology of China, Hefei, China, in 2005 and 2010, respectively. He is currently a Professor with the School of Artificial Intelligence, Optics and Electronics (iOPEN), Northwestern Polytechnical University, Xi'an, China. His research interests include computer vision and remote sensing. For more information, please visit \url{https://crabwq.github.io/}.
\end{IEEEbiography}

\vfill

\end{document}